\documentclass[lettersize,journal]{IEEEtran}

\usepackage{amsmath,amssymb,amsfonts}
\usepackage{algorithmic}
\usepackage{algorithm}
\usepackage{array}
\usepackage{balance}
\usepackage{cases}
\usepackage{cite,hyperref}
\usepackage{color}
\usepackage{graphicx}
\usepackage{mathtools}
\usepackage{stfloats}
\usepackage[caption=false,font=normalsize,labelfont=sf,textfont=sf]{subfig}
\usepackage{tcolorbox}
\usepackage{textcomp}
\usepackage{theorem}
\usepackage{tikz}
\usepackage{url}
\usepackage{verbatim}
\usepackage{xcolor}

\usetikzlibrary{shapes,arrows}
\input{mysymbol.sty}

\usepackage{xr}

\makeatletter
\newcommand*{\addFileDependency}[1]{
  \typeout{(#1)}
  \@addtofilelist{#1}
  \IfFileExists{#1}{}{\typeout{No file #1.}}
}
\makeatother

\newcommand*{\myexternaldocument}[1]{%
    \externaldocument{#1}%
    \addFileDependency{#1.tex}%
    \addFileDependency{#1.aux}%
}

\myexternaldocument{supp}

\newtcolorbox{myblockt}[1]{colback=urblue!5!white,
	colframe=urblue,fonttitle=\bfseries,
	title=#1}
\newtcolorbox{myblock}{colback=urblue!5!white,
	colframe=urblue,fonttitle=\bfseries}

\begin{document}

\title{Joint Network Topology Inference via a Shared Graphon Model}

\author{Madeline Navarro,~\IEEEmembership{Student Member,~IEEE}, and Santiago Segarra,~\IEEEmembership{Senior Member,~IEEE}
\thanks{M. Navarro and S. Segarra are with the Dept. of ECE, Rice University. Preliminary results were presented at ICASSP 2022 \cite{navarro2022graphon}. This work was partially supported by NSF under award CCF-2008555. This work has supplementary material provided by the authors.
Emails:  \href{mailto:nav@rice.edu}{nav@rice.edu}, \href{mailto:segarra@rice.edu}{segarra@rice.edu}}
}



 
\maketitle

\begin{abstract}
We consider the problem of estimating the topology of multiple networks from nodal observations, where these networks are assumed to be drawn from the same (unknown) random graph model.
We adopt a graphon as our random graph model, which is a nonparametric model from which graphs of potentially different sizes can be drawn.
The versatility of graphons allows us to tackle the joint inference problem even for the cases where the graphs to be recovered contain different number of nodes and lack precise alignment across the graphs.
Our solution is based on combining a maximum likelihood penalty with graphon estimation schemes and can be used to augment existing network inference methods. 
{The proposed joint network and graphon estimation is further enhanced with the introduction of a robust method for noisy graph sampling information.}
We validate our proposed approach by comparing its performance against competing methods in synthetic and real-world datasets.
\end{abstract}


\section{Introduction}
Networks conveniently capture systems with complicated relationships and intuitively represent structure via dyadic connections.
Data consisting of entities in interconnected systems, tangible or abstract, are ubiquitous in multiple fields.
Network structures are highly utilized across these many disciplines for representation and analysis of complex information~\cite{brugere2018network}, such as ecology for predicting animal behavior~\cite{farine2015constructing}, neuroscience for modeling relationships between neurons~\cite{narayan2016mixed}, and environmental science for discovering and predicting outcomes of climate relationships~\cite{donges2009complex}.
Interpretation of networks varies greatly depending on the application.
For example, networks may represent physical systems, such as road networks or joint connectivity for skeletal data~\cite{das2020graph}.
In contrast, the nature of a connected system may be more abstract, as is the case for correlation networks, where connectivity represents statistical interdependencies between observed variables. 
Additionally, networks may or may not be directly observable.
As an intuitive example, consider that structural (anatomical) connectivity between neurons can be directly measured~\cite{sporns2013human}, whereas functional connectivity between brain regions cannot be known but must be estimated by observing neural responses~\cite{narayan2016mixed,hutchison2013dynamic}.

While networks are convenient and interpretable tools for tasks on complex data, the underlying structure may be unavailable.
For instance, the true structure of unobservable networks cannot be provided but must be obtained, as with brain functional networks~\cite{narayan2016mixed,hutchison2013dynamic}, the correlations in social behavior between animals~\cite{farine2015constructing}, or any abstract network where we cannot measure connectivity patterns directly.
Alternately, the underlying network may be expensive to obtain, as with brain structural connectivity~\cite{sporns2013human}.
The ubiquitous problem of recovering network connectivity from graph measurements has been well studied in fields such as statistics~\cite{kolaczyk2009statistical} and signal processing~\cite{mateos2019connecting}.
Given data in the form of nodal observations, network connectivity via data-driven methods include graphical models \cite{friedman2008sparse,meinshausen2006highdimensional}, structural equation models \cite{cai2013sparse}, and graph signal processing (GSP)-based approaches \cite{mateos2019connecting,kalofolias2016how,segarra2017network,roddenberry2021network}. 

In the case of inferring the topologies of multiple networks, separate estimation is a feasible methodology.
However, in many scenarios a joint inference method may achieve better performance by leveraging common structures between the graphs to be inferred.
For instance, one would expect certain levels of similarities between the brain networks of different healthy individuals or between the same social network observed at different points in time.
In this paper, we consider the prevalent problem of inferring the topology of multiple networks while assuming that networks share structural similarities.
Many applications rely on multiple instances of interconnected relationships observed over time or in several scenarios, and these complex structures can be conveniently represented by a set of networks.
E.g., Brain networks, functional or structural, are valuable tools for diagnosis, and estimating multiple networks is necessary for analysis of many patients or scenarios~\cite{narayan2016mixed}.
One of the most prominent scenarios requiring the acquisition of multiple networks is when networks vary over time~\cite{hutchison2013dynamic}.
For example, we would expect that the social network of a species of interest will evolve over time~\cite{de2011dynamics}.

We consider the problem of \emph{recovering the connectivity of multiple networks} whose structures are represented by graphs assumed to be \emph{sampled from the same (unknown) random graph model}.
We adopt the nonparametric network model as a graphon~\cite{lovasz2012large}, but the specific graphon model need not be known a priori for our proposed method.
While estimation of multiple networks is well-studied, to the best of our knowledge we provide the only work in the literature that utilizes a shared graphon relationship to jointly estimate graphs of potentially different sizes.

The contributions of our paper are as follows:
\begin{enumerate}
    \item~We present a methodology to {\it infer multiple networks that potentially lack node alignment and may have different sizes} by leveraging the assumption that graphs come from the same nonparametric network model.
    \item~We detail how our approach can be combined with existing network inference methods, effectively providing a whole family of methods to solve the problem of interest.
    \item~We develop a robust version of the problem where the exact sampling criteria of the graphs from the graphon are not known, but we are instead given noisy sampling information.
    \item~Through numerical experiments in synthetic and real-world data we demonstrate the performance of our method in comparison with separate inference and competing joint inference algorithms.
\end{enumerate}

The remainder of this paper is organized as follows. 
We review graph signal processing and graphons in Section~\ref{S:prelim}, along with past related work.
We introduce the problem of interest in Section~\ref{S:prob_statement}.
Our proposed problem formulations are presented in Section~\ref{S:netinf}, and the algorithm development is discussed in Section~\ref{S:alg}. 
We expand the problem to a noisy graph sampling setting and present a robust solution in Section~\ref{S:robust_netinf}.
Section~\ref{S:expmt} presents experimental results of all proposed algorithms on synthetic and real-world data.
Finally, we close with conclusions and discussions of future directions in Section~\ref{S:conc}.

\section{Preliminaries}
\label{S:prelim}

\subsection{Notation}
\label{Ss:notation}
The following notation will be used in this paper. 
We represent vectors as bold lowercase letters $\bbx$, entries of which are indexed by $x_i$.
Matrices are bold uppercase letters $\bbX$, where entries are indexed by $X_{ij}$, and
$\bbX_i$ represents the $i$-th row of $\bbX$.
The superscript $^\top$ denotes the transpose.
A matrix with a calligraphic letter subscript $\bbX_{\ccalI}$ denotes the submatrix of $\bbX$ consisting of rows of $\bbX$ indexed by the set $\ccalI$.
The notation $\bbX_\ccalI^\top$ is ordered by first selecting the rows indexed by $\ccalI$ then transposing the result. 
Thus, selecting the $i$-th column of a matrix $\bbX$ is represented by $[\bbX^\top]^\top_i$ and columns of $\bbX$ indexed by $\ccalI$ is denoted by $[\bbX^\top]_{\ccalI}^\top$.
We define three special index sets $\ccalL$, $\ccalU$, and $\ccalD$ referring to the lower triangle, upper triangle, and diagonal indices of a square matrix. 
A square matrix $\bbX\in\mathbb{R}^{N\times N}$ with the subscript $\ccalL$ as $\bbX_\ccalL$ returns a column vector of length $N(N-1)/2$ of the vertical concatenation of the lower triangular entries of $\bbX$.
The sets $\ccalU$ and $\ccalD$ return similar column vectors.
We let $\bbI_N$ and ${\bf 1}_N$ represent the identity matrix of size $N\times N$ and the all-ones column vector of length $N$.
The Kronecker product, the Kronecker sum, and the Hadamard product are denoted by $\otimes$, $\oplus$, and $\circ$, respectively.
We use $\text{vec}(\bbX)$ to represent the column vector containing the vertical concatenation of the columns in $\bbX$.

\subsection{Graph Signal Processing}
\label{Ss:gsp}

We consider undirected, unweighted graphs of the form $\ccalG=(\ccalV,\ccalE)$ with node (vertex) set $\ccalV$ of cardinality $N$ and edge set $\ccalE\subseteq\ccalV\times\ccalV$. 
The structure of a graph can be represented by its adjacency matrix $\bbS\in\{0,1\}^{N\times N}$, where $S_{ij}\neq0$ if and only if the edge $(i,j)$ exists in the network, and $S_{ij}=0$ otherwise. 
We define graph signals as real-valued observations at each of the $N$ nodes, represented by a vector $\bbx\in\mathbb{R}^N$.
We may associate these nodal values with the graph topology via graph signal models.
Choices for graph signal models include stationary signals that result from diffusion processes over the graph~\cite{sandryhaila2013discrete,segarra2017optimal,marques2017stationary} or as multivariate random numbers, where the graph structure represents statistical dependencies between variables \cite{friedman2008sparse,meinshausen2006highdimensional}.

\begin{figure*}
\centering
	\begin{minipage}[c]{.47\textwidth}
		\includegraphics[width=\textwidth]{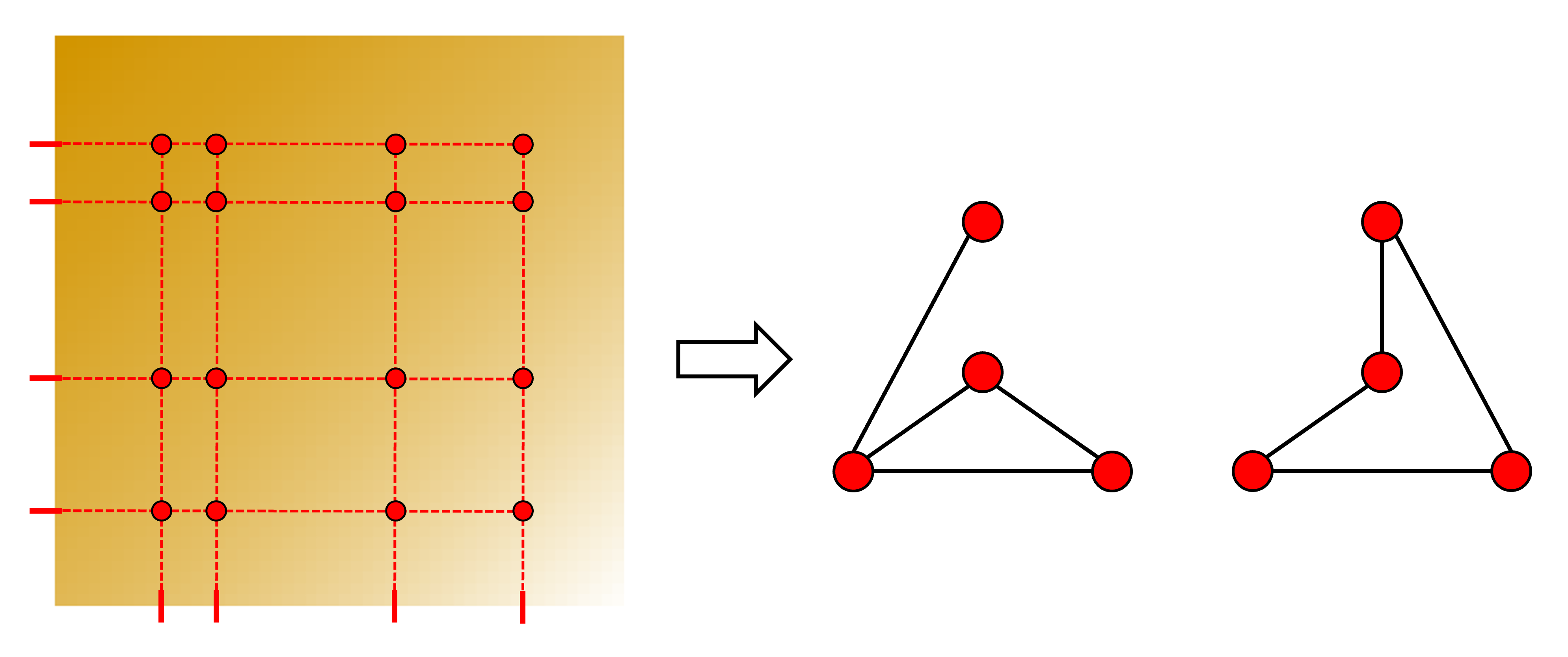}
		
		\centering{\small (a)}
	\end{minipage}
	\hspace{.5cm}
	\begin{minipage}[c]{.47\textwidth}
		\includegraphics[width=\textwidth]{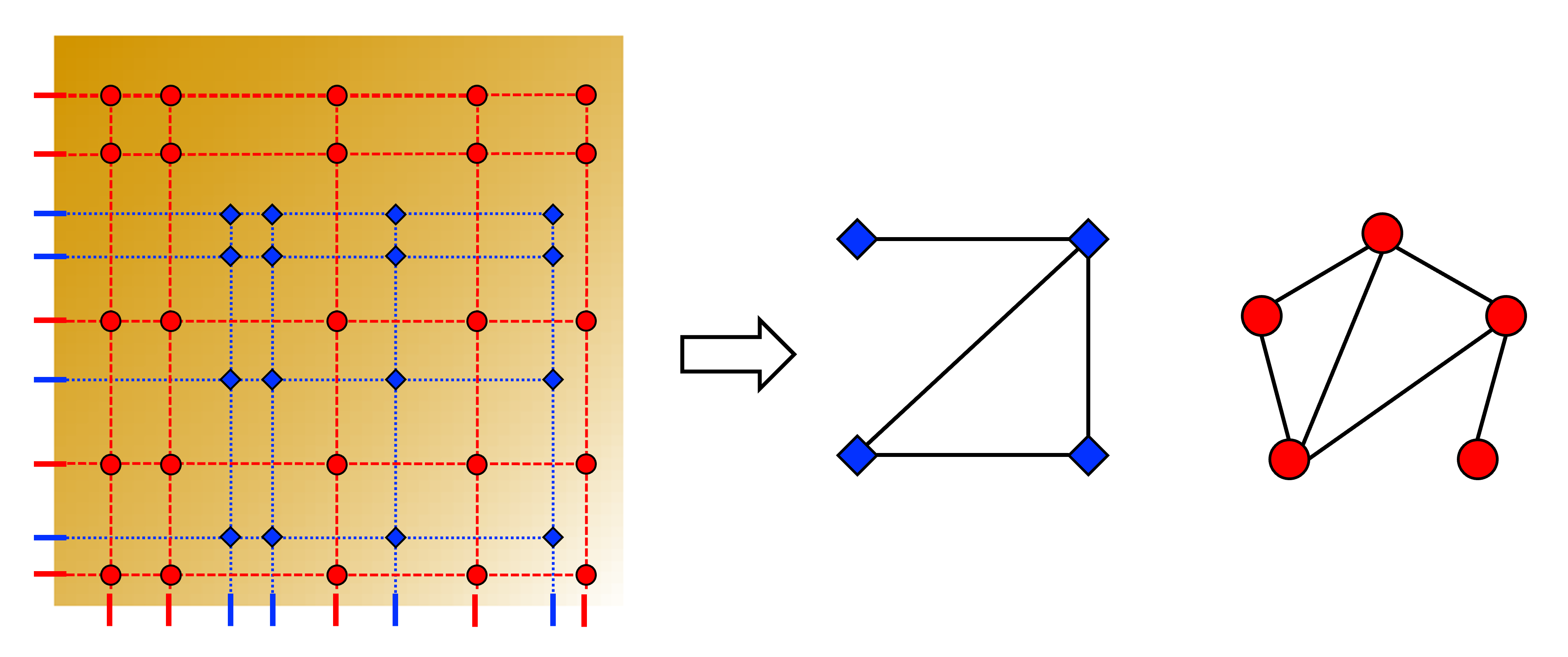}
		
		\centering{\small (b)}
	\end{minipage}
\caption{Schematic depiction of the two problem modalities considered. (a)~Multiple graphs sampled from the {\em same latent point sets} in the {\em same graphon}. Sampled graphs have not only the same size but also node alignment; see Section~\ref{Ss:netinf_probmat}. (b)~Multiple graphs sampled from {\em different latent point sets} in the {\em same graphon}. Sampled graphs may have different sizes; see Section~\ref{Ss:netinf_graphon}. }
	\vspace{-4mm}
\label{fig_cases}
\end{figure*}

\subsection{Graphons}
\label{Ss:graphons}
A graphon is a bounded symmetric measurable function $\ccalW: [0,1]^2\rightarrow[0,1]$ whose domain can be interpreted as edges in an infinitely large adjacency matrix, while the range of $\ccalW$ represents edge probabilities.
By this definition, a graphon can be seen as a random graph model from which graphs with similar structural characteristics can be sampled~\cite{diaconis2007graph,lovasz2012large,avella2020centrality}.
Generating an undirected graph $\ccalG=(\ccalV,\ccalE)$ from a graphon $\ccalW$ consists of two steps: (1)~selecting a random value between 0 and 1 for each node, and (2)~assigning an edge between nodes with probability equal to the value of the graphon at the their randomly sampled points. 
Formally, the steps are as follows
\begin{subequations}
\begin{alignat}{2}
&\zeta_{i} \sim \text{Uniform}([0,1]) &&\forall~i\in\ccalV,
\label{equ_graphon_samp1}\\
& S_{ij} = S_{ji} \sim \text{Bernoulli}\left( \ccalW(\zeta_i,\zeta_j) \right) &\quad& \forall~(i,j)\in\ccalV\times\ccalV,
\label{equ_graphon_samp2}
\end{alignat}
\end{subequations}
where the latent variables $\zeta_i\in[0,1]$ are independently drawn for each node $i$.
This notion of graphon encompasses many commonly used exchangeable distributions on networks.
Indeed, Erd\H{o}s-R\'enyi graph models are represented via constant graphons~\cite{erdHos1959renyi} and stochastic block models (SBMs) via piecewise-constant graphons~\cite{holland1983stochastic}.

In our case, we assume that graphs are sampled from the same graphon, which is also unknown.
Therefore, we propose a method to jointly estimate both the graphs and the underlying graphon.

\subsection{Related Work}
\label{Ss:related}

\noindent{\bf Joint network inference.}
Inferring multiple networks has been well-studied, particularly for graphical model estimation, but most methods require strong assumptions about sizes and node alignment across networks.
Statistical methods that estimate multiple graphical models typically involve modifications of joint graphical lasso with novel penalties encouraging similarity among networks~\cite{wang2020high,gao2016estimation,guo2011joint,danaher2014joint,peeters2020rags2ridges}.
While far fewer than their statistical counterpart, GSP-based methods for inferring multiple networks are prevalent \cite{rey2021joint,araghi2019k,karaaslanli2021multiview,navarro2020joint}.
A particularly prominent scenario of joint network inference is estimation of time-varying graphs, where typical assumptions include slow or smooth network changes over time \cite{hallac2017network,gibberd2017regularized,zhou2010time,yu2022efficient} or smooth graph signals, that is, signal values across nodes vary little \cite{araghi2021dynamic,yamada2021temporal,kalofolias2017learning,yamada2020time}.
A more challenging problem is the inference of multiple networks where it is not known from which network observed data are generated. 
We note GSP methods \cite{araghi2019k,hong2021demixing,ricchi2021dynamics,maretic2020graph} and statistical methods \cite{kao2017disc,hao2017simultaneous,lotsi2016sparse} within this category.
We refer the reader to the reviews in \cite{tsai2021joint,mateos2019connecting,dong2019learning} for more examples of multiple network estimation.
%

The majority of joint network inference methods require network estimation on the same node set (and, thus, of the same size). 
One exception includes brain network estimation with brain regions of different coarseness \cite{pircalabelu2016mixed}, resulting in potentially different sized networks; however, we not only allow for regions at possibly different coarseness levels, but networks may consist of completely different sets of regions altogether.

Furthermore, assumptions on similarity between graph structures mostly concentrate on edgewise relationships, with exceptions including~\cite{arroyo2021inference,soloveychik2017joint,tao2016multiple,mohan2014node} that encourage similar structural characteristics without relying on edgewise comparisons in particular.
Similarly to our paper, \cite{arroyo2021inference}~introduced a random graph model for estimating networks that share a common subspace without encouraging similar edge structures.
Unlike our approach, all of these methods require networks to have the same size.

\vspace{.1in}
\noindent{\bf Graphon estimation.}
The proposed estimation procedure jointly estimates a set of graphs along with their unknown shared graphon. 
Graphon estimation is a well-studied problem \cite{gao2015rate,klopp2019optimal}, where the random graph model is estimated from a binary adjacency matrix.
Methods include estimating the graphon as a continuous smooth object~\cite{sischka2022based,yang2014nonparametric,latouche2016variational,chan2014consistent} along with the coarser SBM estimation~\cite{xu2020learning,wolfe2013nonparametric,olhede2014network,airoldi2013stochastic,choi2012stochastic,rohe2011spectral}.
In many cases, the entire graphon is not needed but only the probability matrix from which the graphon was sampled, i.e., the value of the graphon $\ccalW(x,y)$ only at the latent sample points $(x,y)\in\zeta\times\zeta$, and these methods include neighborhood smoothing \cite{zhao2019change,zhang2017estimating,qin2021iterative}, matrix completion \cite{song2016blind,keshavan2010matrix,borgs2017thy}, or low-rank approximations \cite{chatterjee2015matrix}. 
%
In this work, we present a framework that can leverage any graphon estimation method for network estimation.

While most graphon estimation methods assume availability of only one sampled adjacency matrix, a few works exist that estimate graphon models from multiple graphs \cite{xu2020learning,airoldi2013stochastic,zhao2019change}. 
In our case, we infer a graphon from multiple networks without prior knowledge of the network structure. 
%

\section{Problem Statement}
\label{S:prob_statement}
Leveraging the proposed graphon relationship among networks, we present an algorithm that simultaneously estimates heterogeneous networks and accounts for a common graphon model.
We consider undirected, unweighted graphs without self-loops and sampled from a shared graphon. Formally, consider a set of $K$ different graphs $\{\ccalG^{(k)}\}_{k=1}^K$ where the $k$-th graph has $N^{(k)}$ nodes. 
The set of undirected, unweighted adjacency matrices is represented by the set of adjacency matrices $\{\bbS^{(k)}\}_{k=1}^K$.
Assume also that there is a set of graph signals provided for each graph, represented by $\bbX^{(k)}:=[\bbx_1^{(k)}~\cdots~\bbx_{r_k}^{(k)}]\in\mathbb{R}^{N^{(k)}\times r_k}$, where the $r_k$ columns contain the graph signals corresponding to the $k$-th graph.
We further assume that all graphs are sampled from the same generative model, a graphon $\ccalW$. 
{With some abuse of notation, we let $\bbS$ and $\bbX$ represent the sets of adjacency matrices $\bbS = \{\bbS^{(k)}\}_{k=1}^K$ and graph signals $\bbX = \{\bbX^{(k)}\}_{k=1}^K$, respectively.}
We present our problem as follows.

\vspace{.1in}
\noindent{\bf Problem 1} {\it Given sets of observations $\bbX=\{\bbX^{(k)}\}_{k=1}^K$ for $K$ graphs, find the adjacency matrices $\bbS=\{\bbS^{(k)}\}_{k=1}^K$ under the assumptions that (AS1) all graphs are sampled from the same (unknown) graphon $\ccalW$ and (AS2) the latent point sets $\zeta^{(k)}$ in \eqref{equ_graphon_samp1} for each graph are known.
}
\vspace{.1in}

The first assumption (AS1) creates a relationship among the graphs, and with it we may improve estimation of graphs by jointly inferring the graph structures given their shared relationship.
The second assumption (AS2) eliminates the identifiability problem for graphon estimation, where multiple graphons can lead to the same random graph distribution \cite{diaconis2007graph}.
When all latent point sets are equivalent, i.e., $\zeta^{(k)} = \zeta$ for all graphs $k\in\{1,2,\dots,K\}$, (AS2) is equivalent to the assumption in previous joint network inference methods, where node alignment is present and known for all pairs of graphs. 
However, assuming possibly different known latent point sets is a weaker assumption than that of previous methods, as we do not require node alignment for the graphs.
Furthermore, in Section~\ref{S:robust_netinf} we relax the assumption (AS2) where the latent point sets are not exactly known and only noisy sets are available.

The assumption (AS2) corresponds intuitively to situations of known sensor placement, such as known locations of electrode placement for neural response data collection or known climate regions to be observed.
For example, the brain functional networks of multiple subjects may be measured by considering the same known brain regions or neurons across subjects \cite{narayan2016mixed}.
Inferred graphs may also correspond to statistical interdependence between pairs of variables in a climate data set, where variables are measured at known spatial regions of earth \cite{donges2009complex}.

\section{Graphon-aided Joint Network Estimation}
\label{S:netinf}

In Sections~\ref{Ss:netinf_probmat} and~\ref{Ss:netinf_graphon} we tackle two versions of Problem 1 of increasing difficulty whereas in Section~\ref{Ss:combining_methods} we explain how these solutions can be combined with existing network inference methods.

\subsection{Graphs and Probability Matrix Estimation}
\label{Ss:netinf_probmat}

First consider the case where all graphs are sampled as in \eqref{equ_graphon_samp1} from the same points in the graphon space, that is, $\zeta^{(k)}=\zeta$ for all $k\in\{1,2,\dots,K\}$; see Fig.~\ref{fig_cases}a.
In practice, this case arises, e.g., when using the same sensor placement under multiple trials or experiments.
Since we only consider edge probabilities in the graphon at points $(x,y)\in\zeta\times\zeta$, we need not consider the whole graphon $\ccalW$ but only the probability matrix $\bbT\in[0,1]^{N\times N}$ that contains the edge probabilities at the sampled points.
The graphs $\bbS^{(k)}$ are then sampled from the same probability matrix $\bbT$, so the graphs all must have the same size, that is, $N^{(k)}=N$ for all $k\in\{1,2,\dots,K\}$.

Given $\bbT$, the log-likelihood of a graph $\bbS^{(k)}$ is
\begin{alignat}{2}
&\log\Pr{\bbS^{(k)}|\bbT} = \sum_{i<j} S^{(k)}_{ij}\log(T_{ij}) + (1-S^{(k)}_{ij})\log(1-T_{ij}),
\nonumber
\end{alignat}
where we have leveraged the fact that, given $\bbT$, edges are drawn independently in our graph model [cf.~\eqref{equ_graphon_samp2}].
Furthermore, we can estimate each edge probability $T_{ij}$ by the sample mean of the graph edges. 
Thus, we estimate the probability matrix $\bbT$ as $\frac{1}{K}\sum_{k=1}^K \bbS^{(k)}$.

Recalling the notation for $\bbX$ and $\bbS$ from Problem 1, consider a generic optimization problem to estimate multiple networks that we formalize as
\begin{alignat}{2}
&\min_{\bbS}~~ && f(\bbS,\bbX) + L(\bbS),
\label{equ_netinf}
\end{alignat}
where the objective function $f(\bbS,\bbX)$ estimates graph structures from the observed graph signals, and $L(\bbS)$ is an additional graph penalty or regularizer; in Section~\ref{Ss:combining_methods} we provide common examples for these functions.
To solve our problem at hand, we propose to append the generic formulation in~\eqref{equ_netinf} with a negative log-likelihood penalty to obtain
\begin{alignat}{2}
&\min_{\bbS,\bbT}~~~ &~~~& f(\bbS,\bbX) + L(\bbS) - \sum_{k=1}^K \log\Pr{\bbS^{(k)}|\bbT} 
\nonumber\\
&\text{~~~s.to} && 
\bbS^{(k)}\in\ccalS_A,~~\bbT=\frac{1}{K}\sum_{k=1}^K \bbS^{(k)},
\label{equ_netinf_probmat}
\end{alignat}
where we jointly estimate the graphs and their shared generative probability matrix $\bbT$. 
The new term promoting maximum likelihood encourages edges to be similar based on shared probabilities in aligned edges.
The estimation of the probability matrix entries is included as the sample mean of the edges in the graphs. 
The set $\ccalS_A$ enforces valid binary adjacency matrices, that is,
\be
\ccalS_A = \left\{ \bbS^{(k)} : \bbS^{(k)}=(\bbS^{(k)})^\top,~S^{(k)}_{ii}=0,~S^{(k)}_{ij}\in\{0,1\} \right\},
\ee
where we consider undirected graphs without self-loops and edges that are unweighted.

As mentioned in the problem statement, the assumption $\zeta^{(k)}=\zeta$ for all $k\in\{1,2,\dots,K\}$ is equivalent to node alignment for all graphs.
We relate \eqref{equ_netinf_probmat} to the task of estimating functional networks among the same brain regions of one subject under a set of discrete stimuli, or observing climate variables among the same geographical regions over several time instances.

\subsection{Graphs and Graphon Estimation}
\label{Ss:netinf_graphon}

We now consider the case where each graph is sampled from different latent point sets and graphs may possibly have different sizes, i.e., $\zeta^{(k)}\neq\zeta^{(k')}$ and $N^{(k)}\neq N^{(k')}$ for $k\neq k'$; see~Fig. \ref{fig_cases}b.
Therefore, each graph is sampled from a potentially different probability matrix $\bbT^{(k)}$, which is the value of the graphon at the points $(x,y)\in\zeta^{(k)}\times\zeta^{(k)}$.
{For this case, let the set of probability matrices be represented by $\bbT = \{\bbT^{(k)}\}_{k=1}^K$.}
The probability matrices provide estimates of the graphon at the known latent point pairs, and each graph provides information about the value of its respective probability matrix.

We again build our joint network and graphon inference framework using \eqref{equ_netinf} as our starting point.
We add a new set of terms to encourage maximum likelihood of the graph structures and the graphon model, along with a penalty to incorporate prior graphon information.
Under this setting, we present a general optimization framework to jointly estimate the graphs, the probability matrices, and the graphon as 
\begin{alignat}{2}
&\min_{\scriptsize
\bbS,\bbT,\ccalW
} &~~~& 
f(\bbS,\bbX) + L(\bbS) - \sum_{k=1}^K \log\Pr{\bbS^{(k)}|\bbT^{(k)}} + g(\ccalW)
\nonumber\\
&\text{~~~~s.to} && \bbS^{(k)}\in\ccalS_A,
\nonumber\\
& &&\bbT^{(k)}=h(\bbS^{(k)}),~~~T^{(k)}_{ij} = \ccalW(\zeta_i^{(k)},\zeta_j^{(k)}),
\nonumber\\
& && \ccalW:[0,1]^2\rightarrow [0,1], ~~~\ccalW(x,y)=\ccalW(y,x),
\label{equ_netinf_graphon}
\end{alignat}
where we include the same negative log-likelihood term as in~\eqref{equ_netinf_probmat}, but each graph is associated with a different probability matrix $\bbT^{(k)}$. 
{The function $h(\bbS^{(k)})$ is a probability matrix estimation method that takes an adjacency matrix $\bbS^{(k)}$ as input, such as network histogram or SBM approximations \cite{olhede2014network,chan2014consistent}.}
The third constraint fits the graphon $\ccalW$ at the known latent point pairs to the values of the probability matrices $\bbT^{(k)}$, and the regularization term $g(\ccalW)$ in the objective imposes a prior on the overall graphon structure.
For example, we may apply a thin-plate spline term \cite{duchon1976interpolation} to estimate a smooth graphon assuming that points $\zeta^{(k)}$ are relatively evenly spaced throughout the interval $[0,1]$.
Other interpolation methods may be applied for smooth graphon estimation, such as inverse distance weighting \cite{shepard1968two}.
The suggested potential interpolation terms assume graphon smoothness to estimate the remainder of the graphon.

The assumptions in \eqref{equ_netinf_graphon} are weaker than those in \eqref{equ_netinf_probmat}, thus a wider range of applications are available.
In the example of estimating brain functional networks, functional connectivity of the same subject may be inferred for different sets of brain regions.
Additionally, climate network inference is often based on correlation or mutual inference measures, which decreases with geographic distance \cite{brugere2018network,donges2009complex}.
Thus, separating inference of climate networks into multiple networks of subregions and applying \eqref{equ_netinf_graphon} may be more practical than estimating a single climate network for a large region, as connectivity is expected to be very low for far apart geographical locations.

\subsection{Examples for Network Inference Methods}
\label{Ss:combining_methods}

Up to this point, we have been considering a generic network inference problem in \eqref{equ_netinf}.
Both the formulations in \eqref{equ_netinf_probmat} and \eqref{equ_netinf_graphon} are applicable to existing network inference methods through specific choices of functions $f(\bbS,\bbX)$ and $L(\bbS)$. 
Assumptions required for each signal model are explored in Section~\ref{Ss:assumptions}.

Consider examples for the function $f(\bbS,\bbX)$ that relate the observed graph signals to the structure of the graphs.
Graph signals may be assumed to be smooth over their respective graphs \cite{kalofolias2016how}, and we apply the penalty
\begin{equation}\label{equ_ref_f_smooth}
f(\bbS,\bbX) = \sum_{k=1}^K \|\bbS^{(k)}\circ\bbZ^{(k)}\|_1,
\end{equation}
where $\bbZ^{(k)}_{ij} = \|\bbX^{(k)}_i-\bbX^{(k)}_j\|^2$ as in \cite{kalofolias2016how}.
Alternatively, we may have graph signals that are the diffusion of noise through a graph filter \cite{segarra2017network,mateos2019connecting}.
In this case, we have stationary graph signals, where the signal covariance $\bbC$ commutes with the adjacency matrix $\bbS$. 
Defining sample covariance matrices as $\bbC^{(k)}=\frac{1}{r_k}\bbX^{(k)}(\bbX^{(k)})^\top\in\mathbb{R}^{N^{(k)}\times N^{(k)}}$, we can write
\begin{equation}\label{equ_ref_f_stat}
f(\bbS,\bbX) = \sum_{k=1}^K \|\bbS^{(k)}\bbC^{(k)} - \bbC^{(k)}\bbS^{(k)}\|_F^2.
\end{equation}

In many applications, the graphs of interest are sparse, so it is common to apply a sparsity constraint for each graph \cite{friedman2008sparse,segarra2017network}.
We may apply this with the penalty function $L(\bbS)$ as
\begin{equation}\label{equ_ref_L_sparse}
L(\bbS) = \sum_{k=1}^K \| \mathrm{vec}(\bbS^{(k)})\|_1.
\end{equation}
Note that while the graphon model results in dense graphs as the graph size grows, individually sampled graphs of finite size may be sparse in the sense that the adjacency matrices may contain many zero entries.
For example, an Erd\H{o}s-R{\'e}nyi graph with low edge probability will have a small ratio of edges to pairs of nodes. 
In this case, inclusion of a sparsity promoting penalty would improve network recovery performance.

If, instead of separately inferring each graph, we wish to promote similar sparsity patterns, we may encourage edge similarity between graphs \cite{navarro2020joint,danaher2014joint} as 
\begin{equation}\label{equ_ref_L_sim}
L(\bbS) = \sum_{k<k'} \|\mathrm{vec}( \bbS^{(k)}-\bbS^{(k')} )\|_1,
\end{equation}
which requires graphs that are not only the same size, but are also on the same node set. 
Thus, the regularizer in~\eqref{equ_ref_L_sim} is applicable to our formulation in~\eqref{equ_netinf_probmat} but not to the one in~\eqref{equ_netinf_graphon}.

Combinations of the described examples for $f(\bbS,\bbX)$ and $L(\bbS)$ are common in existing works.
For instance, graph signal stationarity in \eqref{equ_ref_f_stat} and sparsity penalties for each graph via \eqref{equ_ref_L_sparse} are applied in \cite{segarra2017network}.
Moreover, joint inference is performed in \cite{navarro2020joint} by combining \eqref{equ_ref_f_stat} and \eqref{equ_ref_L_sim}.

\subsection{Assumptions}
\label{Ss:assumptions}

The proposed formulations in \eqref{equ_netinf_probmat} and \eqref{equ_netinf_graphon} offer flexible methods for inference of multiple networks, where we may enforce characteristics in the inferred networks based on prior knowledge. 
What follows is an organization of the assumptions required for our approach.

\medskip
\noindent{\bf Problem assumptions.}
Section~\ref{S:prob_statement} introduces the assumptions about our considered problem. 
Assumption (AS1) requires that networks share the same generating graphon.
Typically, joint inference of multiple networks seeks similar edge values by explicitly encouraging edges of the same node pairs to be as close as possible (see Section~\ref{Ss:related}), while our stochastic approach is less stringent. 
For each node pair that occurs in multiple networks, each network shares the \emph{likelihood} that an edge will connect the node pair.
Furthermore, we allow nodes to belong to different node sets across networks.
Given node values in a latent space, edge probabilities for two node pairs are similar if their pairs of latent points are close to each other.
This is conceptually similar to the case of shared node pairs, but now the latent space dictates similarity in stochastic behavior for all edges.

In assumption (AS2), we require knowledge of node assignments in a latent space, which we may incorporate via nodal features whose interactions dictate the presence of edges.
For example, if two nodes belonging to the same class are expected to be connected, and the features of each node indicate its class assignment, then we may apply nodal features to inform the points in the latent space.
We demonstrate an example of using node features for latent point assignment in Section~\ref{Ss:senate_expmt}, where it is more likely for two senators (nodes) to be connected if their political parties (classes) are the same.

\medskip
\noindent{\bf Graph signal model.}
For the penalty \eqref{equ_ref_f_smooth}, we assume that graph signals are smooth on their respective graphs, where we expect well-connected nodes to have similar signal values~\cite{kalofolias2016how}.
In particular, we assume that data on the graphs lie on a smooth manifold, where nodes denote points in the manifold space and edges reflect distances between points.
We apply penalty~\eqref{equ_ref_f_stat} for stationary graph signals, where we assume that graph signals are diffusions of arbitrary input through graph filters. 
If we assume \emph{linear, shift-invariant} graph filters and \emph{white noise} input signals, then the graph signal covariances and the network structures become spectrally related. 
In particular, eigenvectors are equivalent, thus the covariance matrices and adjacency matrices commute. 

\medskip
\noindent{\bf Network and graphon structure.}
We often wish to obtain the most parsimonious network representations for interpretability of connectivity and mitigation of downstream computation. 
It is common to encourage sparsity in inferred networks via the $\ell_1$-norm or Frobenius norm, hence the penalty~\eqref{equ_ref_L_sparse}.
The joint network inference trademark of encouraging edge similarity in \eqref{equ_ref_L_sim} is effective, but it requires that nodes lie on the same node set, an assumption which we relax in \eqref{equ_netinf_graphon}. 
When all networks share the same node set, combining the edge-wise similarity penalty \eqref{equ_ref_L_sim} and shared edge probabilities in our proposed formulation \eqref{equ_netinf_probmat} is an appropriate course of action.

Finally, the choice of $g(\ccalW)$ in~\eqref{equ_netinf_graphon} allows us to incorporate desirable characteristics of the graphon.
For example, we assume smoothness in the underlying graphon, where two node pairs with similar latent point pairs are expected to have similar edge probabilities. 

\section{Algorithm Development}
\label{S:alg}

We address both \eqref{equ_netinf_probmat} and \eqref{equ_netinf_graphon} via an alternating direction method of multipliers (ADMM) algorithm \cite{boyd2011distributed}.
ADMM is an attractive approach as it allows decoupling terms that cannot easily be optimized jointly and handling nonconvex constraints such as our unweighted graph condition.
We terminate optimization after reaching the criteria presented in \cite[Section 3.3.1]{boyd2011distributed}.
We solve \eqref{equ_netinf_graphon} by alternately optimizing the adjacency matrices, the probability matrices, and the shared graphon.
To avoid redundancy, we delay presentation of our solution to \eqref{equ_netinf_probmat}, as it is a simpler version of \eqref{equ_netinf_graphon}.
Indeed, if we let the penalty $g(\ccalW)=0$ and the constraint $\bbT^{(k)} = \bbT = \frac{1}{K}\sum_{k=1}^K \bbS^{(k)}$ for all graphs $k\in\{1,2,\dots,K\}$, then the problem \eqref{equ_netinf_graphon} reduces to the formulation in \eqref{equ_netinf_probmat}.

The general formulation \eqref{equ_netinf_graphon} is difficult to solve due to the continuous graphon penalties and constraints.
However, we can relax it to a computationally feasible problem by estimating a \emph{discretized graphon} \cite{klimm2021modularity,choi2012stochastic,chan2014consistent}.
We replace the graphon $\ccalW$ with a discrete matrix counterpart $\bbW\in[0,1]^{G\times G}$. 
Selection of the size $G$ requires a tradeoff between the fineness of the grid to accurately estimate the graphon $\ccalW$ and the coarseness of the matrix $\bbW$ to minimize computational complexity \cite{klimm2021modularity}.
In this paper, we let the size $G$ be dependent on the network sizes, e.g., $D + \sum_k CN^{(k)}$ for integers $C\geq1$ and $D\geq0$, where values of $C$ and $D$ can be set as large as computational ability allows.
We empirically observe that this choice of discretization is adequate for our proposed joint network and graphon estimation.
We leave investigating other choices of graphon discretization as future work \cite{maclachlan2014theoretical,klimm2021modularity}.

Our proposed relaxation is as follows
\begin{alignat}{2}
&\min_{\scriptsize
\bbS,\bbT,\bbW
} &~~~& 
f(\bbS,\bbX) + L(\bbS) - \sum_{k=1}^K \log\Pr{\bbS^{(k)}|\bbT^{(k)}} + \bar{g}(\bbW)
\nonumber\\
&\text{~~~~s.to} && \bbS^{(k)}\in\ccalS_A,~~\bbW\in\ccalS_{\ccalW},
\nonumber\\
& &&\|\bbT^{(k)}-h(\bbS^{(k)})\|_F^2\leq\epsilon_1^{(k)},
\nonumber\\
& &&\|\bbT^{(k)} - \bbW_{\bbz^{(k)}\bbz^{(k)}}\|_F^2\leq\epsilon_2^{(k)},
\label{equ_netinf_graphon_discr}
\end{alignat}
where $\bbW_{\bbz^{(k)}\bbz^{(k)}}\in[0,1]^{N^{(k)}\times N^{(k)}}$ is the submatrix of $\bbW$ whose entries consist of the graphon $\ccalW$ values at points $(x,y)\in\zeta^{(k)}\times\zeta^{(k)}$ for each graph, and $\bar{g}(\bbW)$ acts as the discretization of the operation $g(\ccalW)$.
Furthermore, the set $\ccalS_{\ccalW}$ defines valid discretized graphon matrices as
\be
\ccalS_{\ccalW} = \left\{ \bbW : \bbW=\bbW^\top, W_{ij}\in[0,1] \right\}.
\ee
{Since we only consider graphs without self-loops, diagonal values of sampled adjacency matrices are ignored~[cf.~\eqref{equ_graphon_samp2}].}

Note that we also relax the equality constraints fitting each probability matrix $\bbT^{(k)}$ to each graphon submatrix $\bbW_{\bbz^{(k)}\bbz^{(k)}}$ and probability matrix estimate $h(\bbS^{(k)})$.
We expect upper bounds $\epsilon_1^{(k)}$ and $\epsilon_2^{(k)}$ to depend on the number of graphs $K$ and the graph sizes $N^{(k)}$.
Greater values of $K$ and $N^{(k)}$ increase the number of sampled points in the graphon space, resulting in more precise graphon estimation.
Additionally, as $N^{(k)}$ grows, the probability matrix $\bbT^{(k)}$ approaches the underlying graphon $\ccalW$, converging to a continuous graphon approximation.

We highlight three major benefits of our formulation: (i)~graphs of different sizes can be inferred, (ii)~explicit knowledge of the graphon $\ccalW$ is not needed, and (iii)~the relaxed problem \eqref{equ_netinf_graphon_discr} is well suited to alternating minimization with optimizing each variable $\bbS$, $\bbT$, and $\bbW$ while fixing the others.
In the sequel, we consider special cases of the functions $f(\bbS,\bbX)$, $L(\bbS)$, $h(\bbS^{(k)})$, and $g(\ccalW)$ to provide a concrete example of our proposed problem and demonstrate its implementation in a common GSP scenario.

\subsection{Stationary Graph Signals and Smooth Graphon}
\label{Ss:special_case}

To demonstrate the implementation of our multiple graph learning algorithm, we present a special case under specific assumptions.
We let $f(\bbS,\bbX)$ take the form of \eqref{equ_ref_f_stat} and let $L(\bbS)=0$.
We choose $h(\bbS^{(k)})$ as a network histogram method to estimate the probability matrices $\bbT^{(k)}$ \cite{chan2014consistent,olhede2014network}.
In particular, the probability matrix corresponding to the adjacency matrix $\bbS^{(k)}$ can be estimated via SBM approximation \cite{chan2014consistent} as
\be
\hat{\bbT}^{(k)} = h(\bbS^{(k)}) = \bbF^{(k)}\bbS^{(k)}\bbF^{(k)},
\ee
where $\bbF^{(k)} = \frac{f}{N^{(k)}-f}(\bbI_{N^{(k)}/f}\otimes ({\bf1}_f{\bf1}_f^\top - \bbI_f))$ computes the empirical edge probability of adjacency matrix blocks with size $f>0$.
We point out that knowledge of the graphon latent sample points obviates the need for sorting the adjacency matrices by degrees before computing the approximate SBM \cite{chan2014consistent}.
Finally, we let $g(\ccalW)$ be a thin-plate spline term \cite{duchon1976interpolation}
\be
g(\ccalW) = \int_0^1\int_0^1 \left(\frac{\partial^2 \ccalW}{\partial x^2}\right)^2 + 2\left(\frac{\partial^2 \ccalW}{\partial x\partial y}\right)^2 + \left(\frac{\partial^2 \ccalW}{\partial y^2}\right)^2 dxdy
\ee
which we then discretize for implementing $\bar{g}(\bbW)$.
First, we introduce the difference matrices $\bbD_1\in\mathbb{R}^{G\times G-1}$ and $\bbD_2\in\mathbb{R}^{G\times G-2}$ such that 
\be
[\bbD_1]_{ij} = 
\begin{Bmatrix}
-1, &~& i-j=0 \\
1, &~& i-j=1 \\
0, &~& \text{otherwise}
\end{Bmatrix}
\ee
and
\be
[\bbD_2]_{ij} = 
\begin{Bmatrix}
1, && i-j\in\{0,2\} \\
-2, && i-j=1 \\
0, && \text{otherwise}
\end{Bmatrix},
\ee
and we discretize $g(\ccalW)$ as 
\be
\bar{g}(\bbW) = \|\bbD_2^\top \bbW\|_F^2 + \|\bbD_1^\top\bbW\bbD_1\|_F^2 + \|\bbW\bbD_2\|_F^2.
\ee

Under these selections, we assume that graph signals are stationary on their respective adjacency matrices; see the discussion of \eqref{equ_ref_f_stat} and Section~\ref{Ss:assumptions}.
Our choices of $h(\bbS^{(k)})$ and $\bar{g}(\bbW)$ assume that the underlying graphon $\ccalW$ is smooth \cite{chan2014consistent,duchon1976interpolation}.

Note that all matrices are symmetric with irrelevant diagonal values.
Thus, we introduce vectors containing the matrix lower triangular entries.
First, let $L^{(k)} = N^{(k)}(N^{(k)}-1)/2$, $J = G(G+1)/2$, and $L_K = \sum_{k=1}^K L^{(k)}$.
We define $\bbs = [(\bbS^{(1)})_\ccalL^\top,\dots,(\bbS^{(K)})_\ccalL^\top]^\top\in\mathbb{R}^{L_K}$, collecting lower triangles of matrices in the tuple $\bbS$ in a column vector.
The vector $\bbt\in\mathbb{R}^{L_K}$ is defined similarly.
We also let $\bbw=[\bbW_{\ccalD}^\top~\bbW_{\ccalL}^\top]^\top$ such that $\bbw\in\mathbb{R}^{J}$ contains all entries of $\text{vec}(\bbW)$ corresponding to the indices $\ccalD\cup\ccalL$.

We introduce the matrix $\bbM$ such that $\|\bbM\bbs\|_2^2 = f(\bbS,\bbX)$ \cite{navarro2020joint}, along with $\bbPsi$ for applying the network histogram method to $\bbs$. We define
\begin{subequations}
\begin{alignat}{2}
&\bbM^{(k)} = [\bbC^{(k)}\oplus-\bbC^{(k)}]_\ccalL^\top + [\bbC^{(k)}\oplus-\bbC^{(k)}]_\ccalU^\top,&
\nonumber\\
&\bbPsi^{(k)} = [\bbF^{(k)}\otimes\bbF^{(k)}]_\ccalL^\top + [\bbF^{(k)}\otimes\bbF^{(k)}]_\ccalU^\top.&
\nonumber
\end{alignat}
\end{subequations}
We also introduce $\bbSigma^{(k)}=[\bbSigma_\ccalD^{(k)}~\bbSigma_\ccalL^{(k)}]$ for selecting graphon indices from $\bbw$ by defining
\begin{subequations}
\begin{alignat}{2}
&\bbSigma_\ccalD^{(k)} = [\bbI_{\bbz^{(k)}}\otimes\bbI_{\bbz^{(k)}}]_\ccalD^\top,&&
\nonumber\\
&\bbSigma_\ccalL^{(k)} = [\bbI_{\bbz^{(k)}}\otimes\bbI_{\bbz^{(k)}}]_\ccalL^\top + [\bbI_{\bbz^{(k)}}\otimes\bbI_{\bbz^{(k)}}]_\ccalU^\top,&&
\nonumber
\end{alignat}
\end{subequations}
where $\bbI_{\bbz^{(k)}}$ denotes a subset of rows of the identity matrix $\bbI_G$ indexed by $\bbz^{(k)}$.
We can then introduce the block matrices $\bbM = \text{blockdiag}(\bbM^{(1)},\bbM^{(2)},\dots,\bbM^{(K)})$, $\bbSigma = [(\bbSigma^{(1)})^\top~(\bbSigma^{(2)})^\top~\cdots~(\bbSigma^{(K)})^\top]^\top$, and $\bbPsi = \text{blockdiag}(\bbPsi^{(1)},\bbPsi^{(2)},\dots,\bbPsi^{(K)})$.
We let 
\be
\bbPhi = \begin{bmatrix}
\bbPhi_{\ccalD,1} & \bbPhi_{\ccalL,1} \\
\bbPhi_{\ccalD,2} & \bbPhi_{\ccalL,2}
\end{bmatrix}
\ee
such that $\|\bbPhi\bbw\|_2^2 = \bar{g}(\bbW)$, where
\begin{subequations}
\begin{alignat}{2}
&\bbPhi_{\ccalD,1} = [\bbD_2\otimes\bbI_G]_\ccalD^\top,&
\nonumber\\
&\bbPhi_{\ccalL,1} = [\bbD_2\otimes\bbI_G]_\ccalL^\top + [\bbD_2\otimes\bbI_G]_\ccalU^\top,&
\nonumber\\
&\bbPhi_{\ccalD,2} = [(\bbD_1\otimes\bbD_1)]_\ccalD^\top,&&
\nonumber\\
&\bbPhi_{\ccalL,2} = [(\bbD_1\otimes\bbD_1)]_\ccalL^\top + [(\bbD_1\otimes\bbD_1)]_\ccalU^\top.&&
\nonumber
\end{alignat}
\end{subequations}
Finally, we define the following function 
\be
\bbGamma(\bbs,\bbt) = -\sum_{i=1}^{L_K} [s_i\log(t_i) + (1-s_i)\log(1-t_i)]
\ee
to represent the log-likelihood term.

From the preceding definitions, we can rewrite the problem \eqref{equ_netinf_graphon_discr} to eliminate constraints and minimize the number of variables to be optimized.
This simplification is shown in the following vectorized problem
\begin{alignat}{2}
&\min_{\scriptsize
\bbs,\bbt,\bbw
} &~~~& 
\frac{\alpha}{2}\|\bbM\bbs\|_2^2 + \bbGamma(\bbs,\bbt) + \frac{\beta}{2}\|\bbPhi\bbw\|_2^2
\nonumber\\
&\text{~~~~s.to} &&
s_i\in\{0,1\},~~w_i\in[0,1],
\nonumber\\
& &&\|\bbt-\bbPsi\bbs\|_2^2\leq \epsilon_1,~~\|\bbt - \bbSigma\bbw\|_2^2 \leq \epsilon_2
\label{equ_netinf_vec}
\end{alignat}
with tuning parameters $\alpha>0$ and $\beta>0$ to control graph signal stationarity and graphon smoothness. 

Developing an ADMM algorithm with guaranteed convergence requires reformulating \eqref{equ_netinf_vec} not only to incorporate dual variables and parameters but also to account for the constraints on the entries in $\bbs$ and $\bbw$.
We introduce the auxiliary variables $\bbp\in\mathbb{R}^{L_K}$ and $\bbv\in\mathbb{R}^J$ and expand the problem as follows
\begin{alignat}{2}
&\min_{\scriptsize
\bbs,\bbt,\bbw
} &~~~& 
\frac{\alpha}{2}\|\bbM\bbs\|_2^2 + \bbGamma(\bbs,\bbt) + \frac{\beta}{2}\|\bbPhi\bbw\|_2^2
\nonumber\\
&\text{~~~~s.to} && \bbs = \bbp,~~\bbw = \bbv,
\nonumber\\
& &&p_i\in\{0,1\},~~v_i\in[0,1],
\nonumber\\
& &&\|\bbt-\bbPsi\bbs\|_2^2\leq \epsilon_1,~~\|\bbt - \bbSigma\bbw\|_2^2 \leq \epsilon_2.
\nonumber
\end{alignat}

The augmented Lagrangian function then takes the form \cite{boyd2011distributed}
\begin{alignat}{2}
&\mathfrak{L}_{\bbrho}&&(\bbs,\bbp,\bbt,\bbw,\bbv,\bbu_1,\bbu_2) = \frac{\alpha}{2}\|\bbM\bbs\|_2^2 + \bbGamma(\bbs,\bbt) + \frac{\beta}{2}\|\bbPhi\bbw\|_2^2
\nonumber\\
& && \quad + \mathbb{I}\Big\{ \bbp_i\in\{0,1\} \forall i\Big\} + \mathbb{I}\Big\{ \bbv_{i}\in[0,1] \forall i \Big\}
\nonumber\\
& && \quad + \rho_1\langle\bbu_1,\bbs-\bbp\rangle + \frac{\rho_1}{2}\|\bbs-\bbp\|_2^2
\nonumber\\
& && \quad + \rho_2\langle\bbu_2,\bbw-\bbv\rangle + \frac{\rho_2}{2}\|\bbw-\bbv\|_2^2
\nonumber\\
& && \quad + \frac{\lambda_1}{2}\|\bbt-\bbPsi\bbs\|_2^2 + \frac{\lambda_2}{2}\|\bbt-\bbSigma\bbw\|_2^2,
\label{equ_netinf_lagrangian}
\end{alignat}
where $\mathbb{I}\{\cdot\}$ is the indicator function that takes the value 0 if the argument is true and infinity otherwise.
The primal variables consist of $\bbs$, $\bbp$, $\bbt$, $\bbw$, and $\bbv$, the dual variables $\bbu_1$ and $\bbu_2$, and the dual parameters $\rho_1$ and $\rho_2$. 
Tuning parameters $\lambda_1$ and $\lambda_2$ determine the strength of the relationships among the adjacency matrices, the graphon, and the probability matrices.
We also define the function $\Pi_\ccalC(\cdot)$ as the projection of the argument onto the set $\ccalC$.

Applying ADMM to \eqref{equ_netinf_lagrangian} results in the following update steps
\begin{subequations}
\begin{alignat}{2}
&\bbs^{j+1} = \argmin_\bbs \mathfrak{L}_{\bbrho} (\bbs,\bbp^{j},\bbt^{j},\bbw^{j},\bbv^{j},\bbu_1^{j},\bbu_2^{j})&
\label{equ_admm_steps_s}\\
&\bbp^{j+1} = \Pi_{\{0,1\}}(\bbs^{j+1} + \bbu_1^{j})&
\label{equ_admm_steps_p}\\
&\bbt^{j+1} = \argmin_\bbt \mathfrak{L}_{\bbrho} (\bbs^{j+1},\bbp^{j+1},\bbt,\bbw^{j},\bbv^{j},\bbu_1^{j},\bbu_2^{j})&
\label{equ_admm_steps_t}\\
&\bbw^{j+1} = \argmin_\bbw \mathfrak{L}_{\bbrho} (\bbs^{j+1},\bbp^{j+1},\bbt^{j+1},\bbw,\bbv^{j},\bbu_1^{j},\bbu_2^{j})&
\label{equ_admm_steps_w}\\
&\bbv^{j+1} = \Pi_{[0,1]}(\bbw^{j+1} + \bbu_2^{j})&
\label{equ_admm_steps_v}\\
&\bbu_1^{j+1} = \bbu_1^{j} + \bbs^{j+1} - \bbp^{j+1}&
\label{equ_admm_steps_u1}\\
&\bbu_2^{j+1} = \bbu_2^{j} + \bbw^{j+1} - \bbv^{j+1}&
\label{equ_admm_steps_u2}
\end{alignat}
\end{subequations}
What follows is a discussion of non-trivial update steps.

\vspace{.1in}
\noindent{\bf Optimization of Graphs.}
Adjacency matrices are initialized via general graph estimation in \eqref{equ_netinf}.
We update the graphs $\bbs$ by solving the problem \eqref{equ_admm_steps_s}, resulting in the closed form solution
\be
(\alpha\bbM^\top\bbM + \rho_1\bbI_{L_K} + \lambda_1\bbPsi^\top\bbPsi)\bbs = \bbgamma(\bbt^j) + \rho_1(\bbp^j-\bbu_1^j) + \lambda_1\bbPsi^\top\bbt^j
\ee
where we let $[\bbgamma(\bbt)]_i = \log(t_i) - \log(1-t_i)$. 

\vspace{.1in}
\noindent{\bf Optimization of Probability Matrices.}
We update the probability matrix vector $\bbt$ by solving
\be
\min_\bbt ~\bbGamma(\bbs,\bbt) + \frac{\lambda_1}{2}\|\bbt-\bbPsi\bbs^{j+1}\|_2^2 + \frac{\lambda_2}{2}\|\bbt-\bbPhi\bbw^j\|_2^2.
\ee
The subproblem for updating $\bbt$ is separable by entries of $\bbt$, thus we can parallelize computation by solving the equivalent problem 
\begin{alignat}{2}
&\min_{t_i} && -(s_i^{j+1}\log(t_i) + (1-s_i^{j+1})\log(1-t_i)) 
\nonumber\\
& && + \frac{c}{2} (t_i - d)^2
\label{equ_netinf_subprob_t}
\end{alignat}
where $c = \lambda_1+\lambda_2$ and $d = \frac{\lambda_1}{c}\bbPsi_i\bbs^{j+1} + \frac{\lambda_2}{c}\bbPhi_i\bbw^{j}$, obtained by completing the square of the original subproblem. 
As the function $s_i\log(t_i) + (1-s_i)\log(1-t_i)$ does not have an easily computed proximal operator, we obtain the update for $\bbt$ by solving \eqref{equ_netinf_subprob_t} via proximal gradient descent (PG) \cite{parikh2014proximal}, and we observe that PG converges quickly in practice.

\vspace{.1in}
\noindent{\bf Optimization of Graphon.}
The solution to the graphon subproblem \eqref{equ_admm_steps_w} is also a closed-form expression,
\be
(\beta\bbSigma^\top\bbSigma + \rho_2\bbI_J + \lambda_2\bbPhi^\top\bbPhi)\bbw = \rho_2(\bbv^j-\bbu_2^j) + \lambda_2\bbPhi^\top\bbt^{j+1}.
\ee

We thus obtain a \emph{convergent result} for solving \eqref{equ_netinf_lagrangian}. 
The main result is shown in the following theorem.

\vspace{.1in}
\noindent{\bf Theorem 1}.\textit{
When ADMM with update steps \eqref{equ_admm_steps_s}-\eqref{equ_admm_steps_u2} is applied to the joint network and graphon inference optimization problem \eqref{equ_netinf_lagrangian}, if the underlying generating graphon $\ccalW$ is bounded away from $0$ and $1$, i.e., there exists some $\epsilon>0$ such that $\ccalW(x,y)\in[\epsilon,1-\epsilon]$ for all $x,y\in[0,1]$, then for large enough parameters $\rho_1$, $\rho_2$, $\lambda_1$, and $\lambda_2$, the resulting sequence $(\bbs^j,\bbp^j,\bbt^j,\bbw^j,\bbv^j,\bbu_1^j,\bbu_2^j)$ has at least one limit point $(\bbs^*,\bbp^*,\bbt^*,\bbw^*,\bbv^*,\bbu_1^*,\bbu_2^*)$, and each is also a stationary point, that is, $0\in\partial\mathfrak{L}_{\bbrho}(\bbs^*,\bbp^*,\bbt^*,\bbw^*,\bbv^*,\bbu_1^*,\bbu_2^*)$. 
}

\vspace{.1in}
\noindent{\bf Proof of Theorem 1.} See Appendix \ref{A:thm1_proof}.

\vspace{.1in}
The most complex ADMM steps are the updates for $\bbs\in\mathbb{R}^{L_K}$ and $\bbw\in\mathbb{R}^J$.
If we assume for simplicity that all graphs have the same number of nodes, $N^{(k)} = N$ for all $k\in\{1,2,\dots,K\}$, then we have that $L_K = KN(N-1)/2$.
We further let the size of the discretized graphon be $G = CKN + D$ for integers $C\geq 1$ and $D\geq 0$.
Thus, each iteration of the update steps \eqref{equ_admm_steps_s}-\eqref{equ_admm_steps_u2} has complexity $\ccalO(J^2) = \ccalO(K^4N^4)$, stemming from the graphon vector $\bbw$ update.

We may precompute the inverses for updating $\bbs$ and $\bbw$, which have complexities $\ccalO(L_K^3) = \ccalO(K^3N^6)$ and $\ccalO(J^3)= \ccalO(K^6N^6)$, respectively.

\vspace{.1in}
\noindent{\bf Remark 1. (Solution to \eqref{equ_netinf_probmat}).}
When solving the problem in \eqref{equ_netinf_probmat}, we let $\bbT^{(k)}=\bbT$ for all $k\in\{1,2,\dots,K\}$ and ignore graphon penalties and constraints, i.e., $\bar{g}(\bbW)$, $\bbW\in\ccalS_{\ccalW}$, and $\| \bbT^{(k)}-\bbW_{\bbz^{(k)}\bbz^{(k)}} \|_F^2\leq\epsilon_2^{(k)}$.
In this case, we forgo update steps \eqref{equ_admm_steps_w}, \eqref{equ_admm_steps_v}, and \eqref{equ_admm_steps_u2}.
The update for $\bbs$ remains the same while letting $\lambda_1=0$. 

For the update of $\bbt$, we note that since $\bbT^{({k})}=\bbT$ for all graphs, we have $L = N(N-1)/2$ degrees of freedom for \eqref{equ_netinf_probmat} instead of the $L_K = \sum_{k=1}^K L^{(k)}$ of \eqref{equ_netinf_graphon}.
Thus we need only estimate one vector $\bbt^{(k)} = \bbt\in\mathbb{R}^{L}$ for all graphs.
We introduce the matrix $\bbR = \frac{1}{K}({\bf 1}_K^\top\otimes\bbI_L)$ such that $\bbt = \bbR\bbs$, equivalent to the last constraint in \eqref{equ_netinf_probmat}.
The update step \eqref{equ_admm_steps_t} then becomes 
\begin{alignat}{2}
&\min_{\scriptsize
t_i
} &~~& 
-K(t_i\log(t_i) + (1-t_i)\log(1-t_i)) 
 + \frac{c'}{2}(t_i-d')^2,
\nonumber
\end{alignat}
where $c' = \lambda_1$ and $d' = \bbR_i\bbs^{j+1}$.
The objective in this subproblem is a difference of convex terms, which can be solved via existing methods such as DCA algorithms \cite{thi2005dc,horst1999dc}.

\vspace{.1in}

\noindent{\bf Remark 2. (Normalized projections).}
The Euclidean projections $\Pi_{\{0,1\}}(\cdot)$ and $\Pi_{[0,1]}(\cdot)$ applied respectively in \eqref{equ_admm_steps_p} and \eqref{equ_admm_steps_v} guarantee convergence of the ADMM algorithm.
However, we empirically observed improved performance by normalized projections.
In particular, the argument $\bbx$ is first normalized as
\be
\bar{\bbx} = \frac{\bbx - \min_i x_i}{\max_j (\bbx - \min_i x_i)},
\ee
and the normalized $\bar{\bbx}$ is then projected to the set $\ccalC$ as usual as $\Pi_{\ccalC}(\bar{\bbx})$. 

\section{Robust Network and Graphon Estimation}
\label{S:robust_netinf}
In practice, the exact placement of the latent sample points may not be available.
For example, electrode location for measuring neural responses may not be precisely comparable across several subjects.
For comparisons of conditions in a climate system, regions of observation may not be consistent over time since atmospheric patterns will not necessarily occur in precisely the same geographical locations.
Thus, we relax the assumption (AS2) that latent points $\zeta^{(k)}$ are exactly known, but approximate points $\bar{\zeta}^{(k)} = \zeta^{(k)} + \omega^{(k)}$ are known, where $\omega^{(k)}$ is random noise perturbing the true sample points.
{This assumption was also applied in~\cite{chandna2021local} for graphon estimation from known graphs.}
We update the problem \eqref{equ_netinf_graphon} as
\begin{alignat}{2}
&\min_{\scriptsize
\bbS,\bbT,\ccalW,\zeta
} &~~~& 
f(\bbS,\bbX) + L(\bbS) - \sum_{k=1}^K \log\Pr{\bbS^{(k)}|\bbT^{(k)}} + g(\ccalW)
\nonumber\\
&\text{~~~~s.to} && \bbS^{(k)}\in\ccalS_A,
\nonumber\\
& &&\bbT^{(k)}=h(\bbS^{(k)}),~~~T^{(k)}_{ij} = \ccalW(\zeta_i^{(k)},\zeta_j^{(k)}),
\nonumber\\
& && \ccalW:[0,1]^2\rightarrow [0,1], ~~~\ccalW(x,y)=\ccalW(y,x),
\nonumber\\
& &&\|\zeta^{(k)} - \bar{\zeta}^{(k)}\|_2^2\leq\epsilon_3^{(k)},
\label{equ_netinf_graphon_robust}
\end{alignat}
where we introduce the final constraint to estimate graphon sample points $\zeta^{(k)}$ based on the given noisy values $\bar{\zeta}^{(k)}$ for all graphs $k\in\{1,2,\dots,K\}$. 

\begin{figure*}
\centering
	\begin{minipage}[c]{.39\textwidth}
		\includegraphics[width=1.1\textwidth]{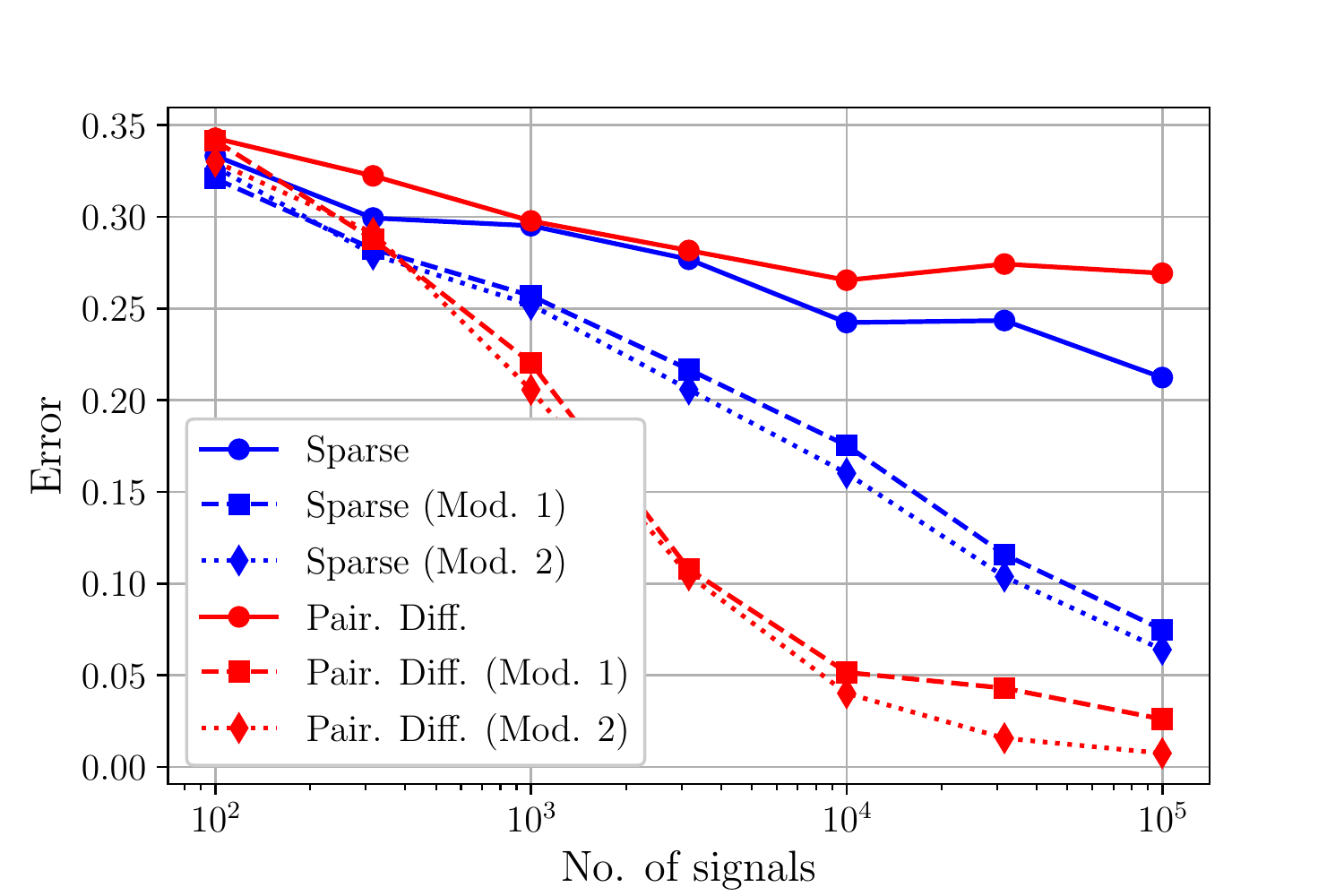}
		\centering{\small (a)}
	\end{minipage}
	\begin{minipage}[c]{.39\textwidth}
		\includegraphics[width=1.1\textwidth]{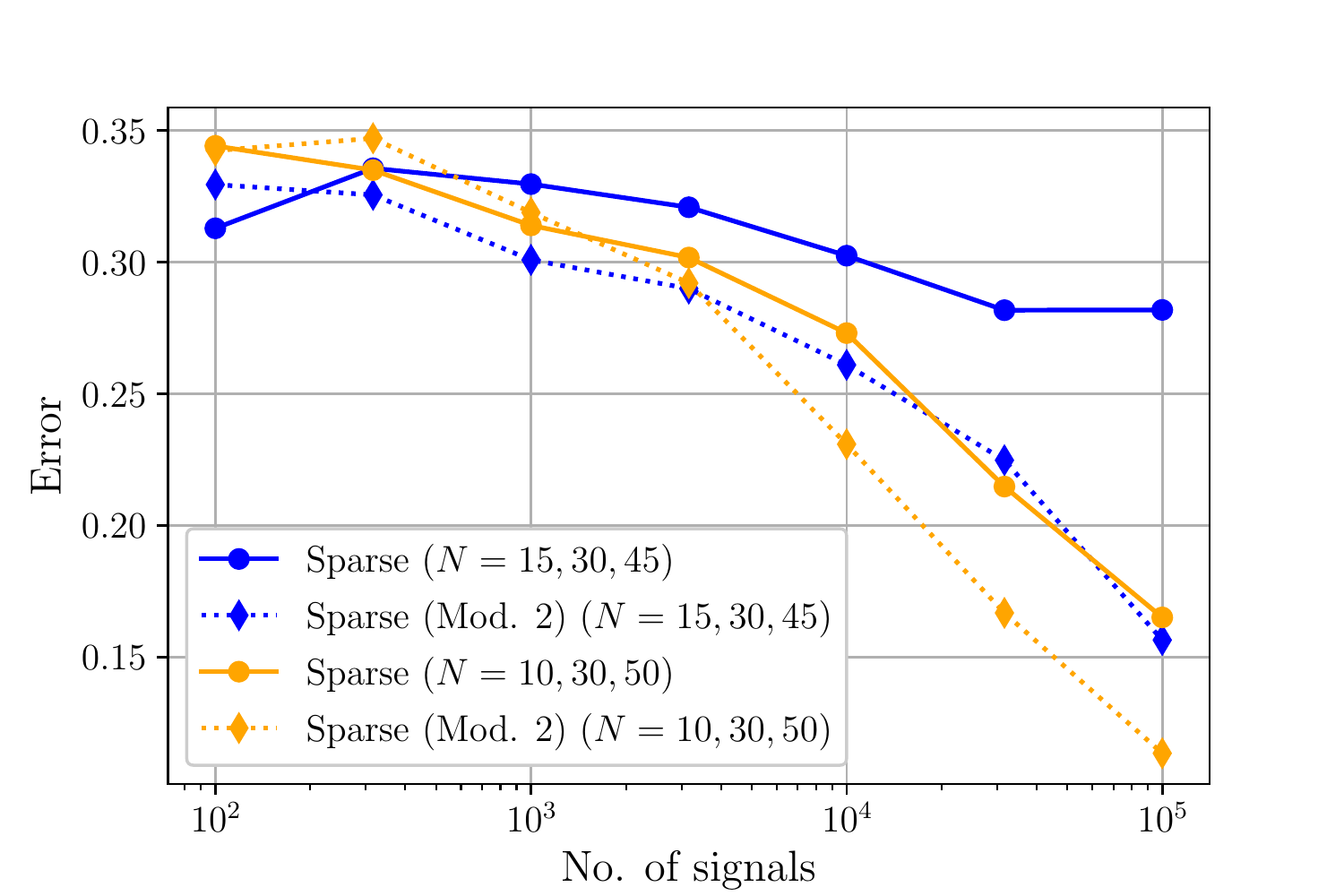}
		\centering{\small (b)}
	\end{minipage}
	\begin{minipage}[c]{.39\textwidth}
		\includegraphics[width=1.1\textwidth]{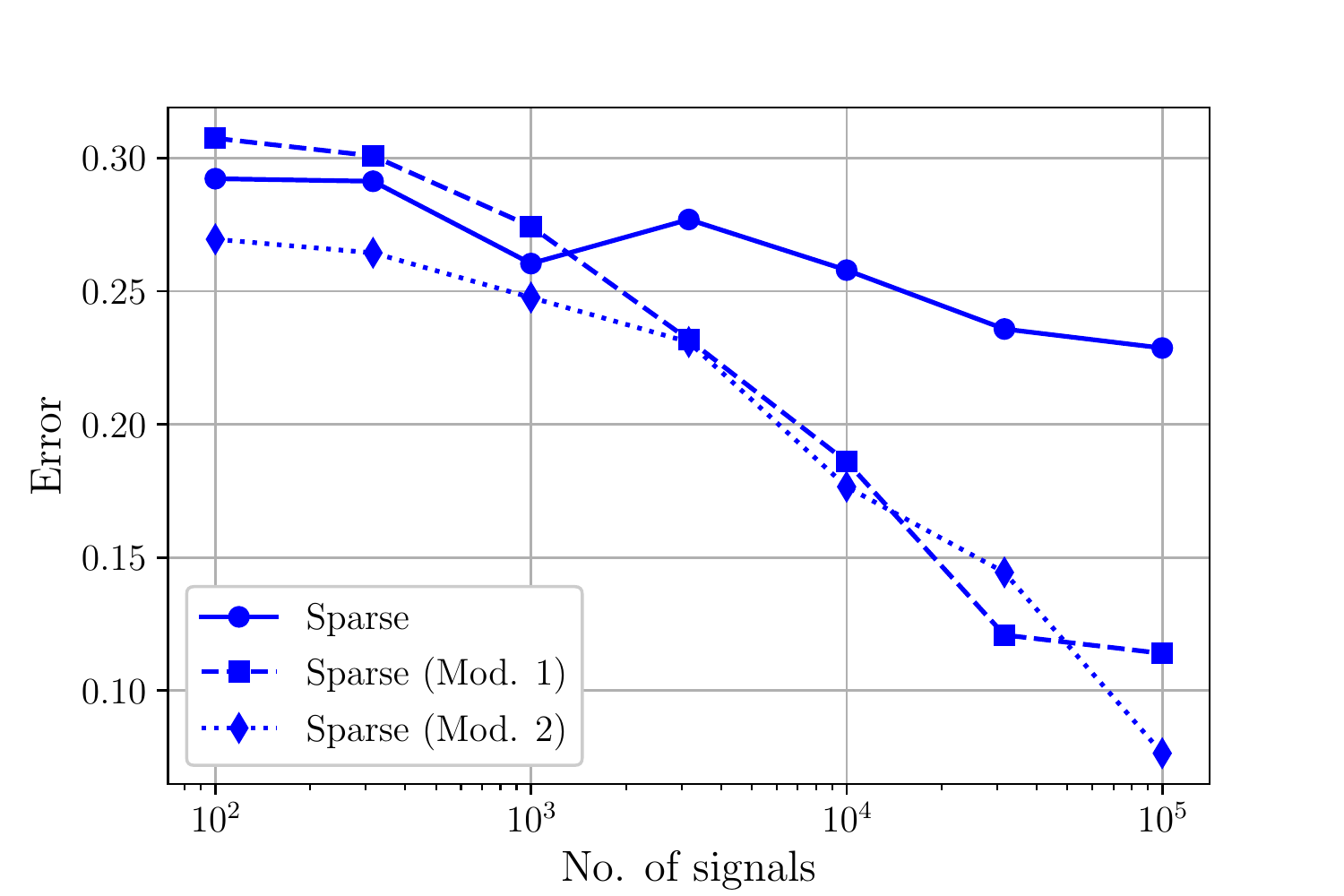}
		\centering{\small (c)}
	\end{minipage}
\caption{ Performance analysis in synthetic networks. (a)~Recovery error for $K=3$ graphs sampled from the {\it same latent point sets} in the same graphon as a function of the number of observed signals. Incorporating the joint estimation of the probability matrix or the graphon both improve estimation performance.~(b)~Recovery error for $K=3$ graphs sampled from {\it latent point sets of different sizes} in the same graphon as a function of the number of observed signals. Separate inference of graphs is outperformed by including joint estimation of the underlying graphon.~(c)~Recovery error for $K=3$ graphs of the same size $N=30$ sampled from {\it different latent point sets} in the same graphon. Joint estimation of networks and the underlying graphon model drastically outperforms separate inference, and the joint network and graphon estimation \eqref{equ_netinf_graphon} demonstrates improvement over joint network and probability matrix estimation \eqref{equ_netinf_probmat}.}
\label{fig_exp}
\end{figure*}

Implementation is almost identical to that of \eqref{equ_netinf_graphon}, where we discretize the graphon and perform an ADMM algorithm. 
The update steps for each variable are the same, with the addition of an update for optimizing the graphon indices $\bbz^{(k)}$.
Let the vector $\bbz$ be the concatenation of the graphon indices, $\bbz = [(\bbz^{(1)})^\top~(\bbz^{(2)})^\top~\cdots~(\bbz^{(K)})^\top]^\top$.
The appended update step solves the subproblem
\begin{alignat}{2}
\bbz^{j+1} = &\argmin_{\bbz} 
&&~
\|\bbt^{j+1} - \bar{\bbSigma}\bbw^{j+1} \|_2^2
\nonumber\\
&\text{s.to} && \!\!\!\! \bar{\bbSigma} = \text{blockdiag}(\bbSigma^{(1)},\bbSigma^{(2)},\dots,\bbSigma^{(K)}),
\nonumber\\
&&&  \!\!\!\! \bbSigma^{(k)} = [(\bbI_{\bbz^{(k)}}\otimes\bbI_{\bbz^{(k)}})^\top_\ccalL]^\top + [(\bbI_{\bbz^{(k)}}\otimes\bbI_{\bbz^{(k)}})^\top_\ccalU]^\top,
\nonumber\\
&&&  \!\!\!\! \bbz^{(k)} = \bar{\bbz}^{(k)} + \Delta\bbz^{(k)},
\nonumber\\
&&&  \!\!\!\! \Delta z^{(k)}_i\in\{-\eta^{(k)},-\eta^{(k)}+1,\dots,\eta^{(k)}\},
\label{equ_admm_steps_z}
\end{alignat}
where $\eta^{(k)}$ is the maximum perturbation of the estimated graphon indices from the given noisy version $\bar{\bbz}^{(k)}$.
The value of $\eta^{(k)}$ depends on the upper bound $\epsilon_3^{(k)}$ and the size $G$ of the discretized graphon.
The area of the error region of radius $\epsilon_3^{(k)}$ in the graphon domain $[0,1]^2$ dictates the size of the error grid of radius $\eta^{(k)}$ in the graphon matrix of size $G\times G$.
We solve this step via a greedy minimization over each entry in $\bbz$ using a grid search over the set of perturbations $\{-\eta^{(k)},-\eta^{(k)}+1,\dots,\eta^{(k)}\}$, where we select the value of $\Delta z^{(k)}_i$ to minimize the objective. 
The optimization order is arbitrary; options include sorted or randomized index orders.

\section{Numerical Experiments}
\label{S:expmt}
\urlstyle{same}

We compare the performance of network topology inference methods with and without the augmentations in \eqref{equ_netinf_probmat}, \eqref{equ_netinf_graphon}, and \eqref{equ_netinf_graphon_robust}, denoted by ``Mod. 1", ``Mod. 2", and ``Mod. 3", respectively.
For all experiments, we apply the same signal model assumption $f(\bbS,\bbX)$ as \eqref{equ_ref_f_stat}, and we compare separate network inference via sparsity penalties \eqref{equ_ref_L_sparse} and joint network inference via pairwise difference penalties \eqref{equ_ref_L_sim}.
For synthetic experiments, we sample from the graphon $\ccalW(x,y)=\frac{1}{2}(x^2+y^2)$.
The error of estimator $\hat{\bbS}$ is calculated as $\|\bbS-\hat{\bbS}\|_F/\|\bbS\|_F$, where the true adjacency matrix is given by $\bbS$.\footnote{Implementations of our method are available at {\url{https://github.com/mn51/jointinf_graphs_graphon}}.}
Additional results are included in the Supplementary Material.

\subsection{Synthetic Experiments}
\label{Ss:synth_expmt}

\vspace{.1in}
\noindent{\bf Same node sets.}
We consider the case where all graphs are sampled from the same points within the graphon space, $\zeta^{(k)}=\zeta$. 
We estimate $K=3$ graphs with $N=30$ nodes as we observe an increasing number of signals for sample covariance computation.
We present in Fig.~\ref{fig_exp}a the comparison of separate and joint network inference methods with the augmentations in \eqref{equ_netinf_probmat} and \eqref{equ_netinf_graphon}, and without either. 
In both methods, the augmented formulations improve estimation performance significantly. 
The pairwise joint penalty \eqref{equ_ref_L_sim} enjoys the greatest improvement, as graphs not only possess node alignment required by \eqref{equ_ref_L_sim}, but they also follow our graph model assumption.

\begin{figure*}
\centering
	\begin{minipage}[c]{.39\textwidth}
		\includegraphics[width=1.1\textwidth]{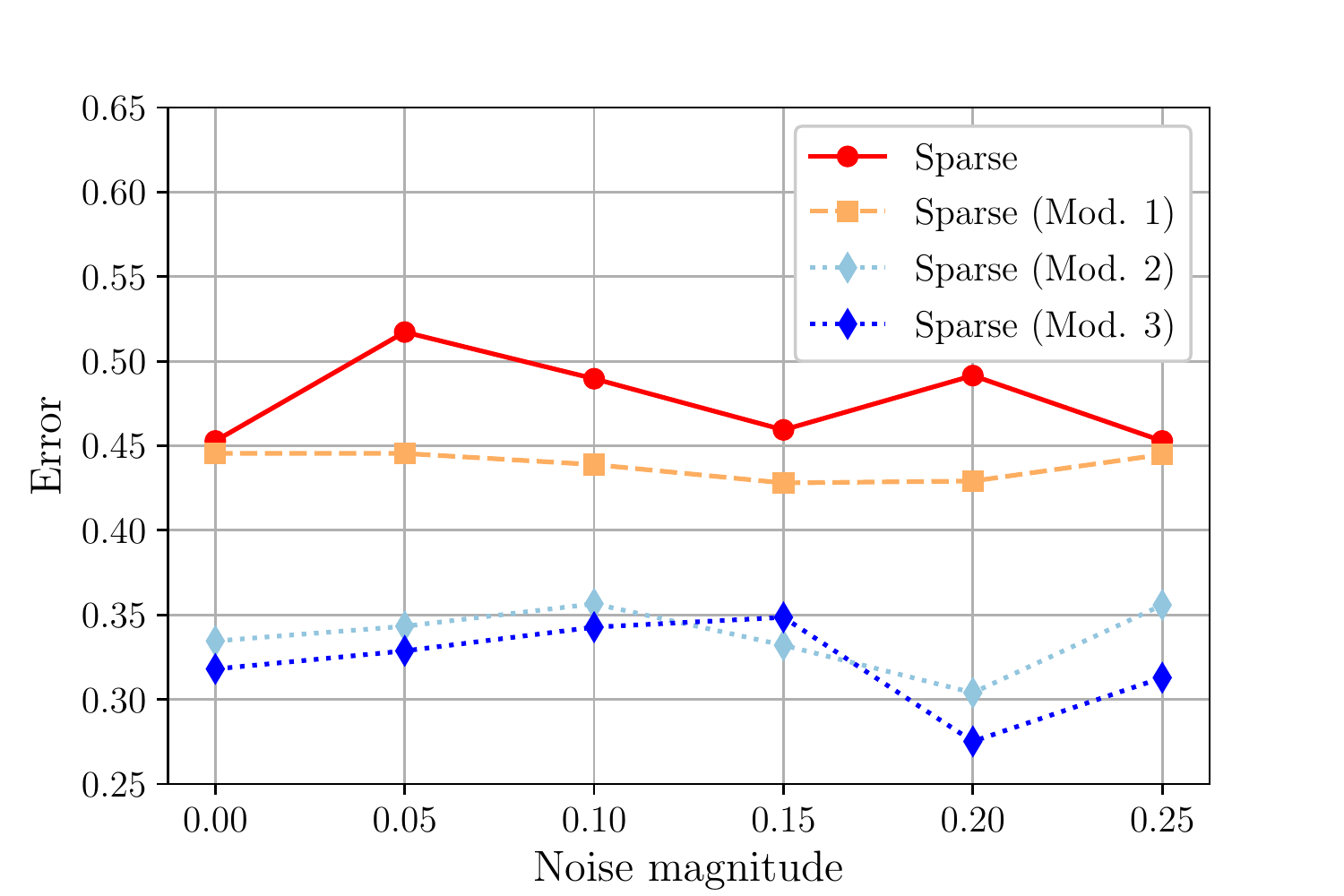}
		
		\centering{\small (a)}
	\end{minipage}
	\begin{minipage}[c]{.39\textwidth}
		\includegraphics[width=1.1\textwidth]{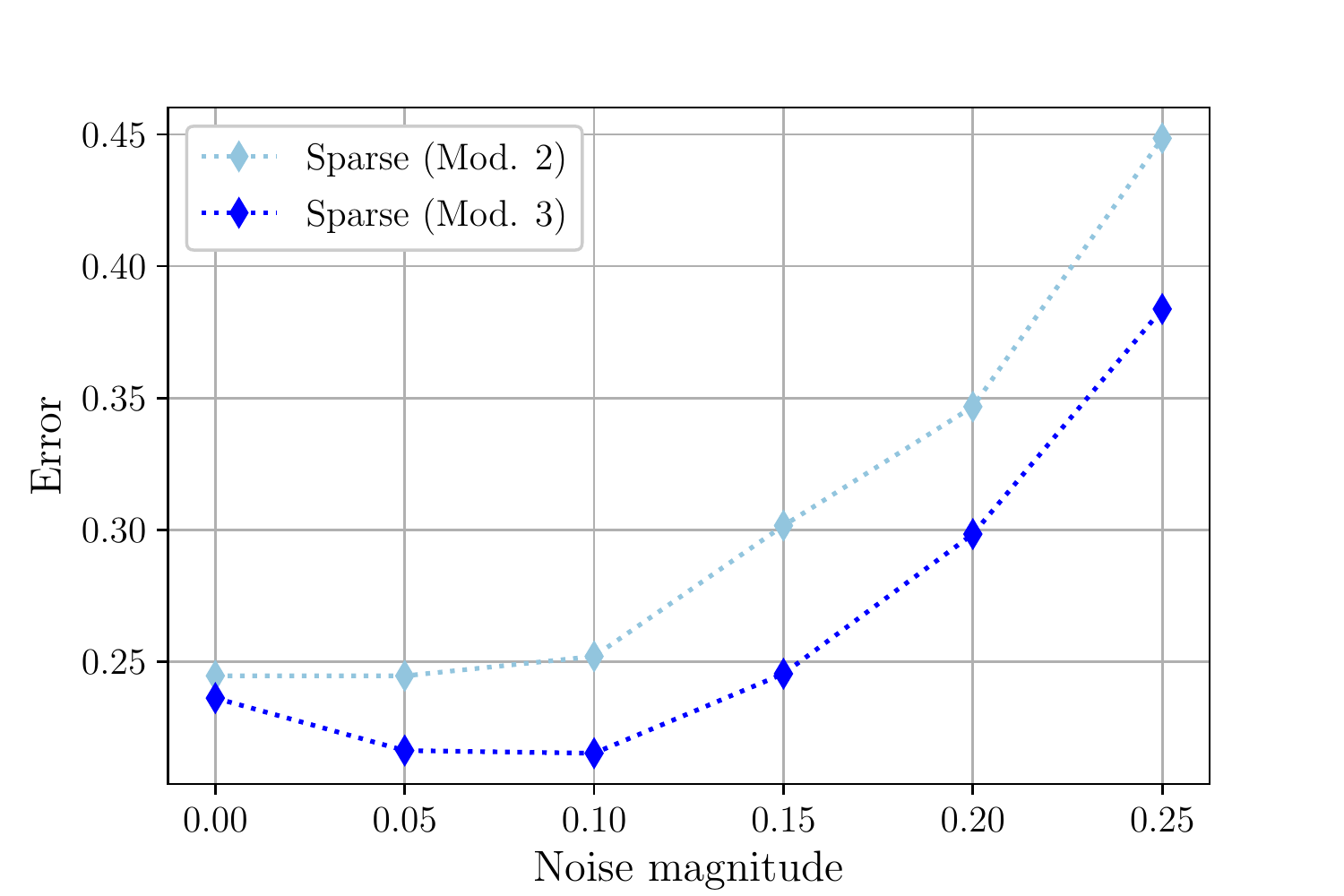}
		
		\centering{\small (b)}
	\end{minipage}

\caption{ Performance analysis in synthetic networks under noisy graphon sample information. (a)~Recovery error of {\it networks} as a function of the magnitude of perturbation of graphon latent sample points $\zeta^{(k)}$ from \eqref{equ_graphon_samp1}. Both versions of joint network and graphon inference outperform separate network inference and joint network and probability matrix inference \eqref{equ_netinf_probmat} for all noise levels. Robust joint inference \eqref{equ_netinf_graphon_robust} exhibits improvement in network recovery compared to \eqref{equ_netinf_graphon}.~(b)~Recovery error of {\it graphon} as a function of the magnitude of perturbation of graphon latent sample points $\zeta^{(k)}$ from \eqref{equ_graphon_samp1}. As the magnitude of perturbation of sample points increases, the robust joint inference \eqref{equ_netinf_graphon_robust} demonstrates increasing recovery performance over \eqref{equ_netinf_graphon}.}
\label{fig_robust}
\end{figure*}

\vspace{.1in}
\noindent{\bf Node sets of different sizes.}
We consider the challenging case where the graphs have different latent point sets of different sizes $N^{(k)}\neq N^{(k')}$.
Unlike the previous experiment, we cannot apply \eqref{equ_netinf_probmat} or \eqref{equ_ref_L_sim}, so we consider only \eqref{equ_ref_f_stat} and \eqref{equ_ref_L_sparse} with and without the joint graphon estimation from \eqref{equ_netinf_graphon}.
We consider $K=3$ graphs for node sets of $N=10,30,50$ and $N=15,30,45$ in Fig.~\ref{fig_exp}b.
For both cases, application of joint graphon inference results in consistent improvement, with increasing performance gap for larger number of observed signals.

\vspace{.1in}
\noindent{\bf Different node sets of same size.}
Finally, we observe graphs of the same size and different node sets, i.e., $N^{(k)}=N^{(k')}$ but $\zeta^{(k)}\neq\zeta^{(k')}$ for every pair $k,k'$. 
In this case, \eqref{equ_netinf_probmat} is applicable, but the model assumption is incorrect, as it assumes that all graphs are not only sampled from the same graphon, but it is also incorrectly assumed that graphs are sampled from the same probability matrix. 
We observe in Fig.~\ref{fig_exp}c the comparison of the three modalities, separate network inference and joint inference with the augmentations in \eqref{equ_netinf_probmat} and \eqref{equ_netinf_graphon}. 
While \eqref{equ_netinf_probmat} outperforms separate inference, indeed \eqref{equ_netinf_graphon} generally exhibits greater improvement as it includes the knowledge of different sample points within the graphon space. 
Our results demonstrate the value in prior knowledge of sampling locations and accurate estimation of the underlying graphon, as both contribute to improvement of network estimation.

\vspace{.1in}
\noindent{\bf Noisy latent sample points.}
Finally, we observe the performance of the robust formulation in \eqref{equ_netinf_graphon_robust} when only a noisy version of the latent sample points $\zeta^{(k)}$ are available. 
We infer $K=3$ graphs of size $N=20$ from the graphon $W(x,y) = 0.25 + 0.75\exp\{-\beta(x-1/2)^2(y-1/2)^2\}$ for $\beta>0$.
We let $\zeta^{(1)},\zeta^{(2)}\sim\text{Unif}(0.4,0.6)$ and $\zeta^{(3)}\sim\text{Unif}(0.2,0.4)$.
Latent sample points are perturbed with increasing levels of magnitude, where $\hat{\zeta}^{(k)} = \zeta^{(k)} + n\omega^{(k)}$ for $n\in\{0,0.05,0.1,0.15,0.2,0.25\}$.
The upper bound $\epsilon_3^{(k)}$ increases in proportion to the magnitude of the noise as $\epsilon_3^{(k)}=n$ for all values of $n$.
Comparisons are shown in Fig.~\ref{fig_robust}. Fig.~\ref{fig_robust}a presents comparisons of graph estimation error for our proposed methods, and Fig.~\ref{fig_robust}b compares graphon estimation error for formulations \eqref{equ_netinf_graphon} and \eqref{equ_netinf_graphon_robust}.
Joint inference of networks and graphons for both the original \eqref{equ_netinf_graphon} and robust \eqref{equ_netinf_graphon_robust} formulations demonstrate consistent superiority in graph recovery over both separate inference and joint inference of networks and probability matrix \eqref{equ_netinf_probmat}.
Furthermore, the robust inference in \eqref{equ_netinf_graphon_robust} demonstrates a general improvement in performance over the original formulation \eqref{equ_netinf_graphon}.
We also observe that graphon estimation is consistently improved when applying the robust joint inference method in Fig.~\ref{fig_robust}b. 
Our approach not only demonstrates a viable method for robust graph estimation under noisy latent sample points but also simultaneously presents improved graphon estimation under perturbed prior information.

\begin{figure*}
\centering
	\begin{minipage}[c]{.39\textwidth}
		\includegraphics[width=1.1\textwidth]{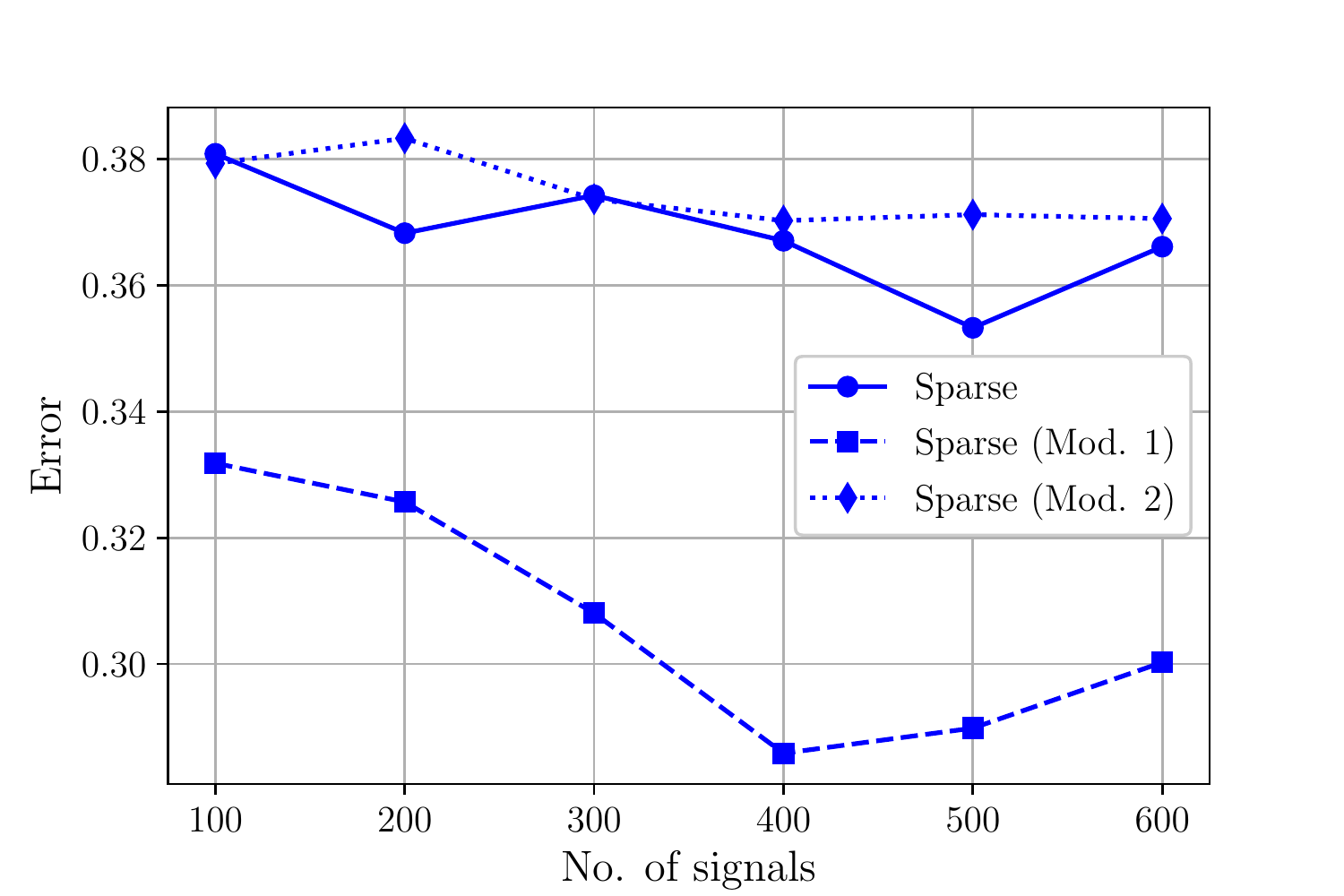}
		
		\centering{\small (a)}
	\end{minipage}
	\begin{minipage}[c]{.39\textwidth}
		\includegraphics[width=1.1\textwidth]{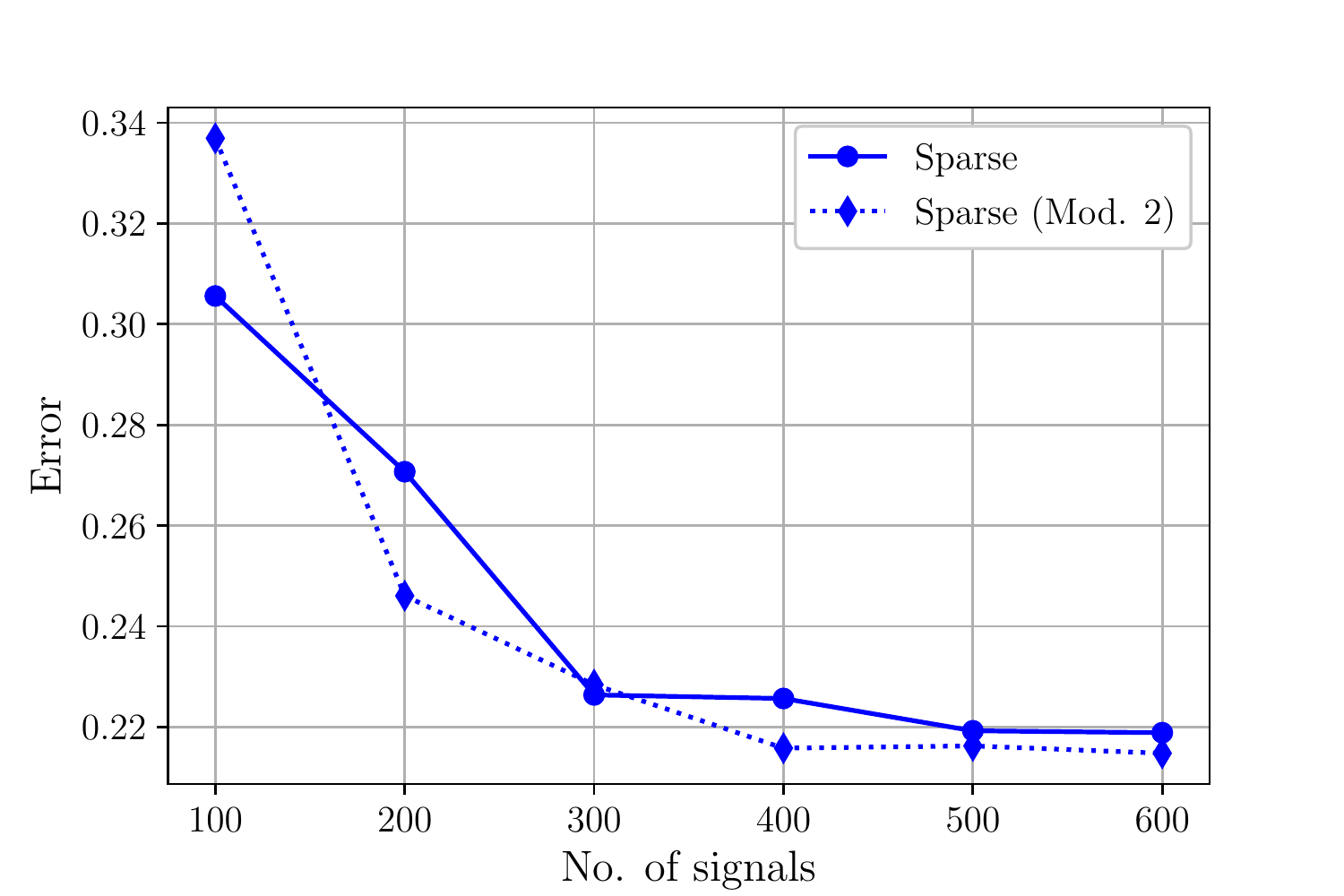}
		
		\centering{\small (b)}
	\end{minipage}
	\begin{minipage}[c]{.39\textwidth}
		\includegraphics[width=1.1\textwidth]{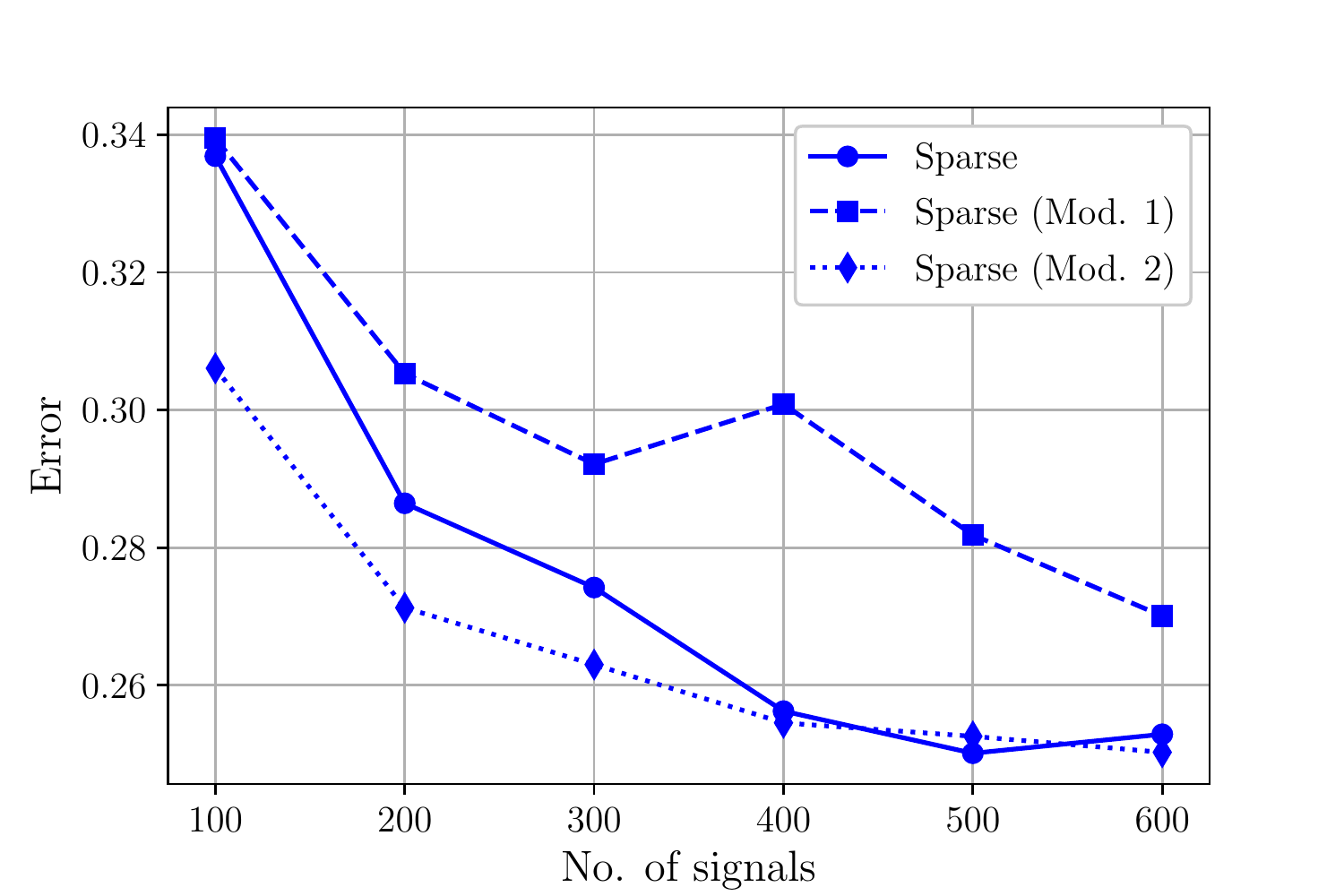}

		\centering{\small (c)}
	\end{minipage}
\caption{Performance analysis in senate networks. (a)~Recovery error as a function of the number of observed signals for {\it induced subgraphs} of three senate networks of the same set of senators. Joint network and probability matrix inference \eqref{equ_netinf_probmat} outperforms both separate network inference and joint network and graphon inference \eqref{equ_netinf_graphon} for all sets of observed signals.~(b)~Recovery error of {\it induced subgraphs} of three senate networks of sizes $N=15,30,45$ as a function of the number of observed signals. Joint network and graphon inference \eqref{equ_netinf_graphon} outperforms separate network inference as the set of observed signals grows larger.~(c)~Recovery error as a function of the number of observed signals for {\it induced subgraphs} of three senate networks of size $N=30$ consisting of different sets of senators. Joint network inference outperforms separate network inference consistently for all sets of observed signals, and joint network and graphon inference \eqref{equ_netinf_graphon} demonstrates better performance than joint network and probability matrix inference \eqref{equ_netinf_probmat}.}
\label{fig_senate}
\end{figure*}

\subsection{Senate Networks}
\label{Ss:senate_expmt}
Finally, we performed graph estimation with real-world data of U.S. congress roll-call votes \cite{lewis2020voteview}, and we set up senate vote signals as in \cite{navarro2020joint}.
The number of nodes corresponds to the number of votes (100 senators and 1 President).
We let votes represent graph signals, where a node can take the value 1 for \textit{yea}, -1 for \textit{nay}, and 0 for abstinence. 
We observe the $724$, $919$, and $612$ votes of congresses 103, 104, and 105, respectively, and we let the underlying true networks be obtained by separate estimation of each network using all available votes. 
For the experiment, we consider the same three cases of network estimation as in the synthetic experiments: (i) networks sampled from the same node set, (ii) networks of different sizes, and (iii) networks of the same size but sampled from different node sets. 
We estimate subgraphs of the separately inferred true graphs, and senators are chosen for each subgraph analogous to sampling a graphon at selected points.

\vspace{.1in}
\noindent{\bf Same set of senators.}
In the first case, we estimate $K=3$ subgraphs sampled from the same points, i.e., networks with nodes $N=30$ consisting of the same $30$ senators, where we only observe votes of senators corresponding to these subsets of nodes. 
We show in Fig. \ref{fig_senate}a that \eqref{equ_netinf_probmat}, which assumes all three node sets consist of the same set of senators, indeed outperforms separate estimation and the augmentation in \eqref{equ_netinf_graphon}, demonstrating the statistical voting similarities of these chosen senate seats across congresses.

\vspace{.1in}
\noindent{\bf Sets of senators of different sizes.}
For networks of different sizes, we estimate induced subgraphs of sizes $N=15,30,45$. 
In Fig. \ref{fig_senate}b we observe that joint network and graphon estimation from \eqref{equ_netinf_graphon} appears to consistently rival or outperform separate inference for larger numbers of observed votes, even though the true networks were estimated separately. 
Indeed, while separate inference of the induced subgraphs becomes more similar to the true network generation as the number of signals increases, \eqref{equ_netinf_graphon} is still able to improve accuracy.

\vspace{.1in}
\noindent{\bf Different senator sets of same size.}
We also revisit the comparison under an incorrect model assumption for \eqref{equ_netinf_probmat}. 
We consider $K=3$ subgraphs of size $N=30$ for all, but the estimated subgraphs consist of different sets of senators. 
Fig. \ref{fig_senate}c demonstrates that estimation via \eqref{equ_netinf_graphon} drastically outperforms that of \eqref{equ_netinf_probmat}, which assumes the subgraphs consist of the same senators, and furthermore \eqref{equ_netinf_graphon} generally outperforms separate inference. 
These experiments demonstrate that the versatile nonparametric nature of graphons can aid the recovery of real-world graphs, even if these graphs have not been explicitly drawn from a graphon model in the first place.

\begin{figure}[!t]
    \centering
        \includegraphics[width=.3\textwidth]{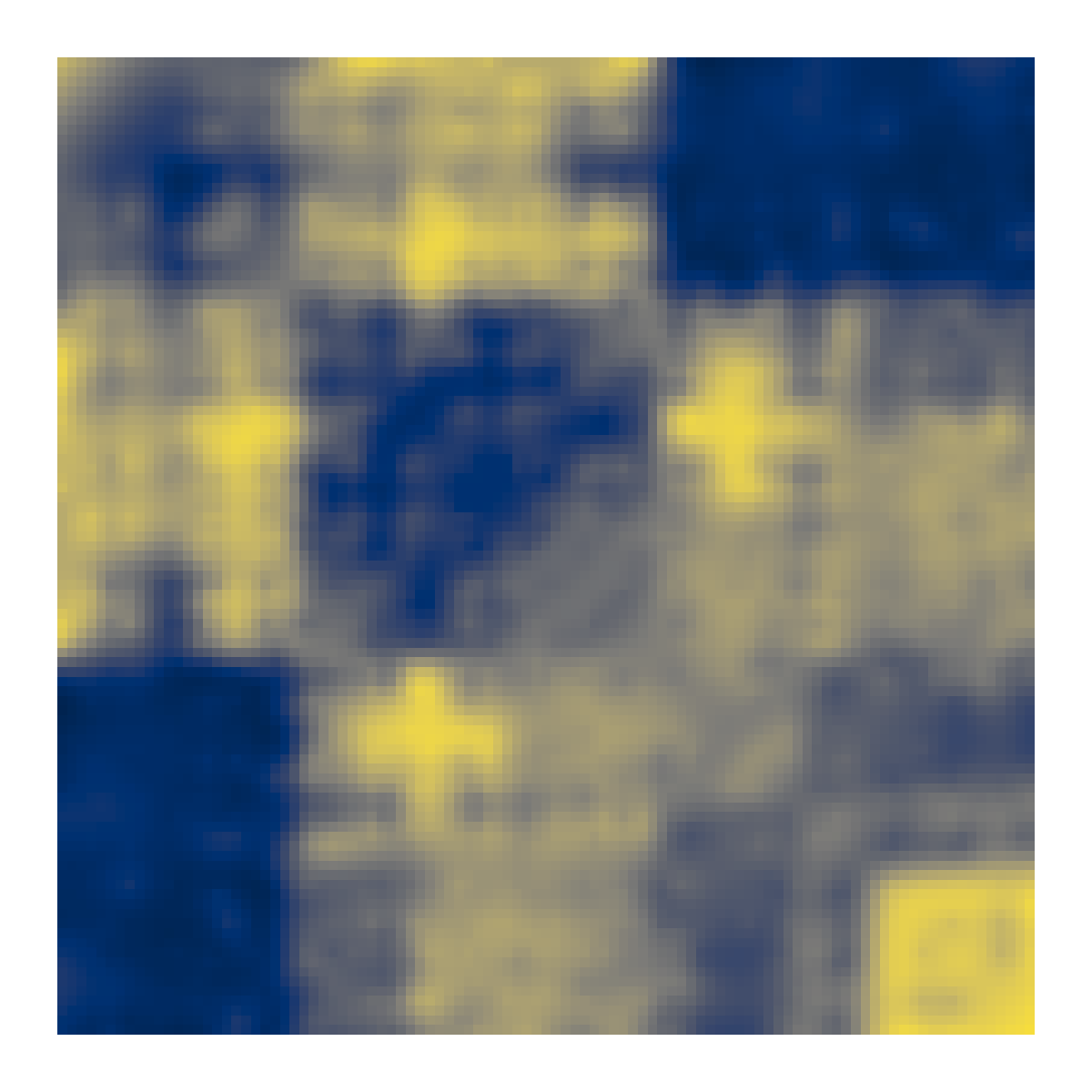}
    \caption{\small{The graphon estimated jointly with senate networks exhibits two-community node clustering behavior.}}
    \label{fig:senate_sbm}
    \vspace{-2mm}
\end{figure}

Finally, we include in Fig.~\ref{fig:senate_sbm} the estimated graphon from applying \eqref{equ_netinf_graphon} for inferring senate networks. 
We assume an SBM-like underlying model, where we expect that the nodes (senators) will exhibit clustering behavior into two communities (political parties). 
The expected behavior is clearly recognizable in the estimated graphon, which validates our claim that the estimated graphon exhibits the general shared structure of a set of real-world networks.

\begin{figure}[!b]
    \centering
        \includegraphics[width=.47\textwidth]{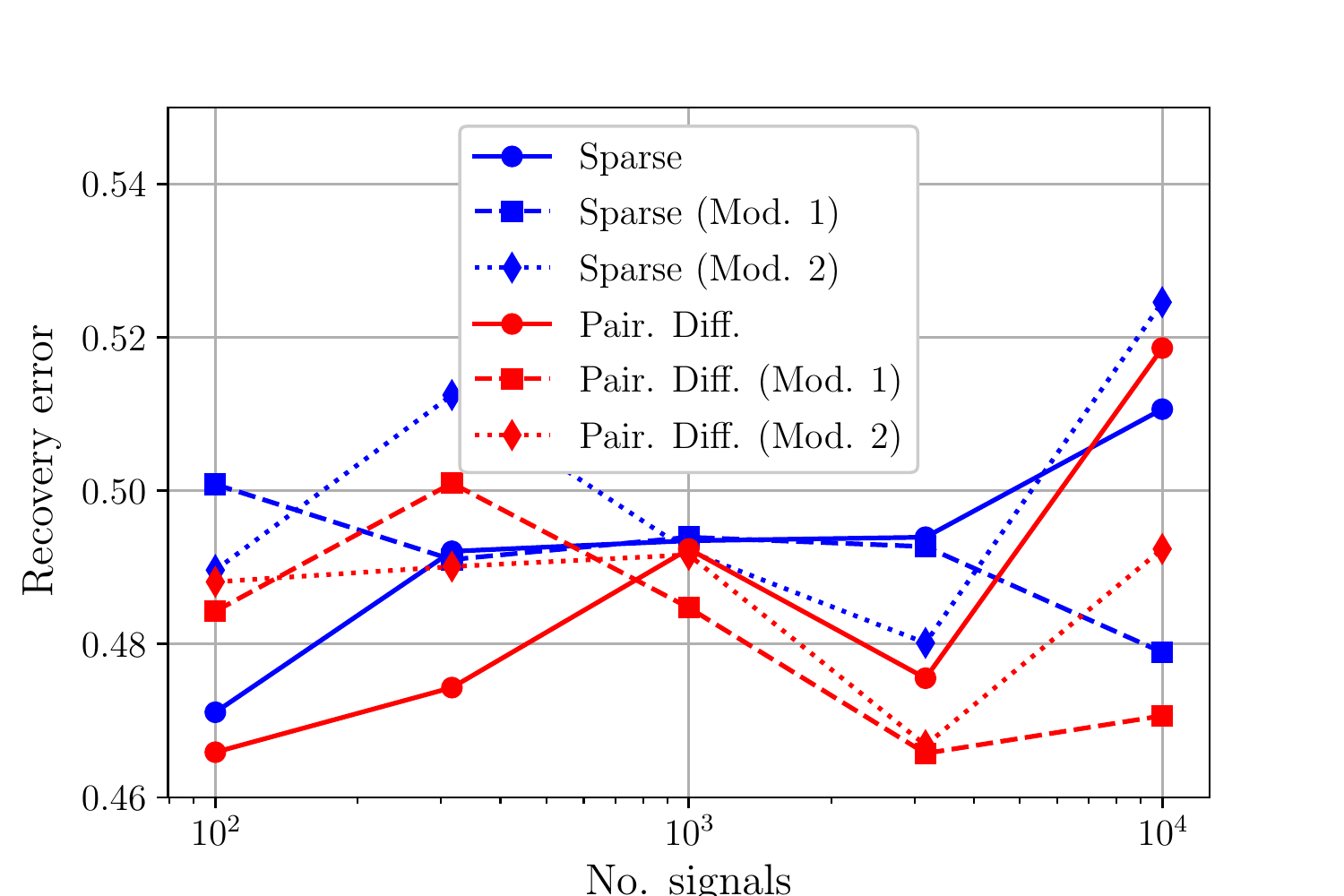}
    \caption{\small{Performance analysis in brain functional networks. Recovery error as a function of observe time frames for the induced subnetworks of $K=3$ different subsets of neurons, each of size $N=30$. 
    Joint network inference assuming an underlying graphon model demonstrates competitive and promising improvement in performance.}}
    \label{fig:s_neural_error}
\end{figure}

\subsection{Brain Functional Networks}
\label{Ss:brain_expmt}

We consider the estimation of neural connectivity using data from the Allen Brain Atlas~\cite{lein2007genome}, which consists of recordings of \emph{in vivo} neural responses from 100 neurons of a live mouse being shown visual stimuli, which are designed to stimulate the visual cortex.
We let the ground-truth brain functional network be the single estimated network given all 100000 available frames of neural data.

Consider the case where we only observe the activity of multiple subsets of neurons of size $N=30$ each, where each subset is measured separately, but we wish to infer the induced subnetworks connecting the observed neurons.
We estimate these brain functional subnetworks via the separate network inference method used to generate the ground-truth network, denoted by ``Sparse", which we compare to our proposed joint inference augmentations.
Estimation results are shown in Fig.~\ref{fig:s_neural_error}, where we observe that for different subsets of neurons, applying our joint inference approaches results in competitive and even superior performances, even when the true underlying network is obtained from the Sparse method.

\section{Conclusion and Future Work}
\label{S:conc}

We demonstrated a method to jointly estimate multiple networks under the assumption that they are sampled from the same graphon. 
To the best of our knowledge, our work is the first to leverage graphons to solve the challenging problem of inferring graphs of different sizes.
We also presented a robust method when only noisy information about the latent variables $\zeta^{(k)}$ is available.
We demonstrated that our proposed maximum-likelihood-based method improves network estimation in synthetic and real-world experiments, along with the efficacy of our robust solution with inexact sampling data.
In terms of future directions, we plan to consider:
i)~Alternative random graph models (beyond graphons) that can also generate graphs of different sizes while promoting other common structural characteristics, and
ii)~Additional relevant tasks on multiple networks that benefit from additional statistical knowledge, i.e., the shared graphon assumption; tasks include graph clustering, dynamic graph change-point detection, or graph node alignment for graphs of different sizes.



{\appendices
\section{Proof of Theorem 1}\label{A:thm1_proof}

This proof is inspired by \cite{wang2019global}, with digression mainly due to the function $g(\bbs,\bbt,\bbw)$, whose Lipschitz smoothness depends on combinations of $\bbs$, $\bbt$, and $\bbw$, and the order of updates, where the lower semicontinuous indicator functions are not updated first.
Additionally, we have that the function $f(\bbp,\bbv)$ is not convex with respect to $\bbp$.

\vspace{.1in}
\noindent{\bf Remark 3. (Boundedness of $\bbt$).}
Note that since the graphon is bounded away from 0 and 1, values of $\ccalW$ and $\bbT^{(k)}$ are restricted to the set $[\epsilon,1-\epsilon]$.
In this proof we assume that $\bbt^j\in[\epsilon,1-\epsilon]$ for all $j\in\mathbb{N}$. 
This can be easily enforced in the optimization by modifying the log-likelihood penalty $\hat{\bbGamma}(\bbs,\bbt) = -\sum_{i=1}^{L_K}[s_i\log(t_i-\epsilon)+(1-s_i)\log(1-t_i-\epsilon)]$, where the resulting modified probability matrix update in \eqref{equ_netinf_subprob_t} remains solvable by proximal gradient descent. 
\vspace{.1in}

We first introduce the following Lemmas to be used in the remainder of the proof. 
Proofs of the Lemmas can be found in the Supplementary Material.

\medskip
\noindent{\bf Lemma 1 (Coercivity).} \textit{The objective function $\phi(\bbs,\bbp,\bbt,\bbw,\bbv)$ is coercive over the feasible set $\ccalF:=\{(\bbs,\bbp,\bbt,\bbw,\bbv)\in\textnormal{dom}(\phi):\bbs=\bbp,~\bbw=\bbv\}$. 
Specifically, $\phi(\bbs,\bbp,\bbt,\bbw,\bbv)\rightarrow\infty$ when $\|(\bbs,\bbp,\bbt,\bbw,\bbv)\|\rightarrow\infty$}.

\medskip

\noindent{\bf Lemma 2 (Objective regularity).} \textit{
If $\bbt\in[\epsilon,1-\epsilon]^{L_K}$ for some $\epsilon>0$, then
\bi[(a)] For positive constants $G_s^s$, $G_s^t$, $G_w^w$, and $G_w^t$, \ei
\begin{flalign}&
\quad\|\nabla_{\bbs} g(\bbs,\bbt,\bbw)-\nabla_{\bbs} g(\hat{\bbs},\hat{\bbt},\hat{\bbw})\|_2^2 
&\nonumber\\&
\quad\qquad \leq \lambda_1^2G_s^s\|\bbs-\hat{\bbs}\|_2^2 
+ \lambda_1^2G_s^t\|\bbt-\hat{\bbt}\|_2^2,
&\nonumber
\end{flalign}
\begin{flalign}&
\quad\|\nabla_{\bbt} g(\bbs,\bbt,\bbw)-\nabla_{\bbt} g(\bbs,\bbt,\hat{\bbw})\|_2^2 
\leq\lambda_2^2G_w^t\|\bbw-\hat{\bbw}\|_2^2,
&\nonumber
\end{flalign}
\begin{flalign}&
\quad\|\nabla_{\bbw} g(\bbs,\bbt,\bbw)-\nabla_{\bbw} g(\hat{\bbs},\hat{\bbt},\hat{\bbw})\|_2^2 
&\nonumber\\&
\quad\qquad\leq \lambda_2^2G_w^w\|\bbw-\hat{\bbw}\|_2^2 + \lambda_2^2G_w^t\|\bbt-\hat{\bbt}\|_2^2 .
&\nonumber
\end{flalign}
\bi[(b)] For positive constants $H_s$ and $H_w$, \ei
\begin{flalign}&
\qquad\|\nabla_{\bbs} h(\bbs,\bbw)-\nabla_{\bbs} h(\hat{\bbs},\hat{\bbw})\|_2^2 
\leq \alpha^2H_s\|\bbs-\hat{\bbs}\|_2^2,
&\nonumber
\end{flalign}
\begin{flalign}&
\qquad\|\nabla_{\bbw} h(\bbs,\bbw)-\nabla_{\bbw} h(\hat{\bbs},\hat{\bbw})\|_2^2 
\leq \beta^2H_w\|\bbw-\hat{\bbw}\|_2^2.
&\nonumber
\end{flalign}
}

\medskip

\noindent{\bf Lemma 3 (Bound dual by primal).} \textit{
For all $j\in\mathbb{N}$, 
\begin{flalign}&
\text{\textnormal{(a)~}}\bbu_1^j = -\frac{1}{\rho_1}\left( \nabla_{\bbs}h(\bbs^j,\bbw^{j-1}) + \nabla_{\bbs}g(\bbs^j,\bbt^{j-1},\bbw^{j-1}) \right) 
&\nonumber\\&
\qquad\qquad + \bbp^{j-1}-\bbp^j
&\nonumber
\end{flalign}
\begin{flalign}&
\text{\textnormal{(b)~}}\bbu_2^j = -\frac{1}{\rho_2}\left( \nabla_{\bbw}h(\bbs^j,\bbw^j) + \nabla_{\bbw}g(\bbs^j,\bbt^j,\bbw^j) \right)
&\nonumber\\&
\qquad\qquad + \bbv^{j-1}-\bbv^j
&\nonumber
\end{flalign}
\begin{flalign}&
\text{\textnormal{(c)~}}\|\bbu_1^j - \bbu_1^{j+1}\|_2^2 \leq \frac{1}{\rho_1^2} \big( (\lambda_1^2G_s^s+\alpha^2H_s)\|\bbs^j-\bbs^{j+1}\|_2^2 \big. 
&\nonumber\\&
\qquad\qquad\qquad\qquad + (1/\epsilon^4+\lambda_1^2G_s^t)\|\bbt^{j-1}-\bbt^j\|_2^2
&\nonumber\\&
\qquad\qquad\qquad\qquad \big. + \rho_1^2\|\bbp^{j-1}-\bbp^j\|_2^2 + \rho_1^2\|\bbp^j-\bbp^{j+1}\|_2^2\big)
&\nonumber
\end{flalign}
\begin{flalign}&
\text{\textnormal{(d)~}}\|\bbu_2^j - \bbu_2^{j+1}\|_2^2 \leq \frac{1}{\rho_2^2} \big((\lambda_2^2G_w^w+\beta^2H_w)\|\bbw^j-\bbw^{j+1}\|_2^2 \big.
&\nonumber\\&
\qquad\qquad\qquad\qquad + \lambda_2^2G_w^t\|\bbt^j-\bbt^{j+1}\|_2^2
&\nonumber\\&
\qquad\qquad\qquad\qquad \big. + \rho_2^2\|\bbv^{j-1}-\bbv^j\|_2^2 + \rho_2^2\|\bbv^j-\bbv^{j+1}\|_2^2 \big)
&\nonumber
\end{flalign}
}

\medskip

We next list four key properties under which ADMM convergence can be ensured.
The Supplementary Material contains proofs that the ADMM steps \eqref{equ_admm_steps_s}-\eqref{equ_admm_steps_u2} satisfy the following properties when applied to problem \eqref{equ_netinf_lagrangian}.

\medskip
\noindent{\bf Property P1 (Monotone, lower-bounded $\mathfrak{L}_{\bbrho}$ and bounded sequence).}\textit{
The sequence $\{\bbs^j,\bbp^j,\bbt^j,\bbw^j,\bbv^j,\bbu_1^j,\bbu_2^j\}$ is bounded, and the Lagrangian $\mathfrak{L}_{\bbrho}(\bbs^j,\bbp^j,\bbt^j,\bbw^j,\bbv^j,\bbu_1^j,\bbu_2^j)$ is lower bounded.}

\medskip

\noindent{\bf Property 2 (Sufficient descent).}\textit{
There is a constant $C_1(\bbrho,\bblambda)>0$ such that for all $k$, we have that
\begin{alignat}{2}&
\mathfrak{L}_{\bbrho}(\bbs^0,\bbp^0,\bbt^0,\bbw^0,\bbv^0,\bbu_1^0,\bbu_2^0)
&\nonumber\\&\qquad\qquad
-\mathfrak{L}_{\bbrho}(\bbs^j,\bbp^j,\bbt^j,\bbw^j,\bbv^j,\bbu_1^j,\bbu_2^j)
&\nonumber\\&\qquad
\geq C_1(\bbrho,\bblambda)\sum_{i=0}^j (
   \|\bbs^i-\bbs^{i+1}\|_2^2
 + \|\bbp^i-\bbp^{i+1}\|_2^2
&\nonumber\\&\qquad\qquad
 + \|\bbt^i-\bbt^{i+1}\|_2^2
 + \|\bbw^i-\bbw^{i+1}\|_2^2
 + \|\bbv^i-\bbv^{i+1}\|_2^2
).
&\nonumber
\end{alignat}
}

\medskip

\noindent{\bf Property 3 (Subgradient bound).}\textit{
There is a constant $C_2(\bbrho,\bblambda)>0$ and subdifferential 
\be \bbd^{j+1}\in\partial\mathfrak{L}_{\bbrho}(\bbs^{j+1},\bbp^{j+1},\bbt^{j+1},\bbw^{j+1},\bbv^{j+1},\bbu_1^{j+1},\bbu_2^{j+1})\ee
such that
\begin{alignat}{3}&
\|\bbd^{j+1}\|_2^2 \leq C_2(\bbrho,\bblambda) (
   \|\bbs^j-\bbs^{j+1}\|_2^2
 + \|\bbp^j-\bbp^{j+1}\|_2^2
&\nonumber\\&\qquad~~
 + \|\bbp^{j-1}-\bbp^j\|_2^2
 + \|\bbt^j-\bbt^{j+1}\|_2^2
 + \|\bbt^{j-1}-\bbt^j\|_2^2
&\nonumber\\&\qquad~~
 + \|\bbw^j-\bbw^{j+1}\|_2^2
 + \|\bbv^j-\bbv^{j+1}\|_2^2
 + \|\bbv^{j-1}-\bbv^j\|_2^2
).
&\nonumber
\end{alignat}
}

\medskip

\noindent{\bf Property 4 (Limiting continuity).}\textit{
If $(\bbs^*,\bbp^*,\bbt^*,\bbw^*,\bbv^*,\bbu_1^*,\bbu_2^*)$ is the limit point of the subsequence $(\bbs^{j_a},\bbp^{j_a},\bbt^{j_a},\bbw^{j_a},\bbv^{j_a},\bbu_1^{j_a},\bbu_2^{j_a})$ for $a\in\mathbb{N}$, then $\mathfrak{L}_{\bbrho}(\bbs^*,\bbp^*,\bbt^*,\bbw^*,\bbv^*,\bbu_1^*,\bbu_2^*)$ is the limit of $\mathfrak{L}_{\bbrho}(\bbs^{j_a},\bbp^{j_a},\bbt^{j_a},\bbw^{j_a},\bbv^{j_a},\bbu_1^{j_a},\bbu_2^{j_a})$ as $a\rightarrow\infty$.
}

\medskip

Finally, under the above four properties, we can demonstrate the convergence of ADMM applied to problem \eqref{equ_netinf_lagrangian}.
By Property 1, we have that the sequence $(\bbs^j,\bbp^j,\bbt^j,\bbw^j,\bbv^j,\bbu_1^j,\bbu_2^j)$ is bounded and thus has a convergent subsequence $(\bbs^{j_a},\bbp^{j_a},\bbt^{j_a},\bbw^{j_a},\bbv^{j_a},\bbu_1^{j_a},\bbu_2^{j_a})$ for $a\in\mathbb{N}$ with the limit point $(\bbs^*,\bbp^*,\bbt^*,\bbw^*,\bbv^*,\bbu_1^*,\bbu_2^*)$ as $a\rightarrow\infty$.
Under Properties 1 and 2, the Lagrangian $\mathfrak{L}_{\bbrho}(\bbs^j,\bbp^j,\bbt^j,\bbw^j,\bbv^j,\bbu_1^j,\bbu_2^j)$ is bounded above and below, so we have that $\|\bbs^j-\bbs^{j+1}\|_2^2\rightarrow0$, $\|\bbp^j-\bbp^{j+1}\|_2^2\rightarrow0$, $\|\bbt^j-\bbt^{j+1}\|_2^2\rightarrow0$, $\|\bbw^j-\bbw^{j+1}\|_2^2\rightarrow0$, $\|\bbv^j-\bbv^{j+1}\|_2^2\rightarrow0$ as $j\rightarrow\infty$.
Then, by Lemma 3 and the primal variable convergence, we also have that $\|\bbu_1^j-\bbu_1^{j+1}\|_2^2\rightarrow0$ and $\|\bbu_2^j-\bbu_2^{j+1}\|_2^2\rightarrow0$ as $j\rightarrow\infty$.
Based on Property 3, there exists a subgradient sequence $\bbd^j\in\partial\mathfrak{L}_{\bbrho}(\bbs^j,\bbp^j,\bbt^j,\bbw^j,\bbv^j,\bbu_1^j,\bbu_2^j)$ such that $\|\bbd^j\|_2^2\rightarrow0$ as $j\rightarrow\infty$.
Thus, we also have that $\|\bbd^{j_a}\|_2^2\rightarrow0$ as $a\rightarrow\infty$.
Finally, by Property 4, we have that $\mathfrak{L}_{\bbrho}(\bbs^*,\bbp^*,\bbt^*,\bbw^*,\bbv^*,\bbu_1^*,\bbu_2^*)$ is the limit of $\mathfrak{L}_{\bbrho}(\bbs^{j_a},\bbp^{j_a},\bbt^{j_a},\bbw^{j_a},\bbv^{j_a},\bbu_1^{j_a},\bbu_2^{j_a})$ as $a\rightarrow\infty$.
By the definition of the general subgradient, we have that $0\in\partial\mathfrak{L}_{\bbrho}(\bbs^*,\bbp^*,\bbt^*,\bbw^*,\bbv^*,\bbu_1^*,\bbu_2^*)$.

}

\bibliographystyle{IEEEtran}
\bibliography{citations}



\vfill

\end{document}


\title{Joint Network Topology Inference via a Shared Graphon Model \textendash~Supplementary Material}

\author{Madeline Navarro,~\IEEEmembership{Student Member,~IEEE}, and Santiago Segarra,~\IEEEmembership{Senior Member,~IEEE}
}
\maketitle

\setcounter{equation}{18}
\setcounter{figure}{6}

\section{Proofs of Theoretical Results}

\noindent{\bf Proof of Lemma 1.} 
If $|\bbs_i|\rightarrow\infty$ ($|\bbw_i|\rightarrow\infty$), then $|\bbp_i|\rightarrow\infty$ ($|\bbv_i|\rightarrow\infty$) and $\bbp_i\notin\{0,1\}$ ($\bbv_i\notin[0,1]$), so $f(\bbp,\bbv)\rightarrow\infty$.
Thus, $\phi(\bbs,\bbp,\bbt,\bbw,\bbv)\rightarrow\infty$ when $\|(\bbs,\bbp,\bbt,\bbw,\bbv)\|\rightarrow\infty$.
For $\bbt\in\text{dom}(\phi)$, we have that $\bbt\in[\epsilon,1-\epsilon]^{L_K}$.
Then, we conclude that $\phi(\bbs,\bbp,\bbt,\bbw,\bbv)$ is coercive over the feasible set $\ccalF$. \hfill$\blacksquare$

\medskip

\noindent{\bf Proof of Lemma 2.}
We have the following bound for the gradient of $\bbGamma(\bbs,\bbt)$ with respect to $s_i$ as
\begin{alignat}{2}&
\left| \log\left(\frac{t_i}{1-t_i}\right) - \log\left(\frac{\hat{t}_i}{1-\hat{t}_i}\right) \right|^2 = 
\left| \log\left(\frac{t_i(1-\hat{t}_i)}{\hat{t}_i(1-t_i)}\right) \right|^2
&\nonumber\\&
\qquad\qquad\qquad\leq \left| \frac{t_i(1-\hat{t}_i)}{\hat{t}_i(1-t_i)} - 1 \right|^2
\leq \frac{1}{\hat{t}_i^2(1-t_i)^2}|t_i-\hat{t}_i|^2
&\nonumber\\&
\qquad\qquad\qquad\leq \frac{1}{\epsilon^4}|t_i-\hat{t}_i|^2.
&\nonumber
\end{alignat}
We can then bound the difference of the gradient $\nabla_{\bbs} g(\bbs,\bbt,\bbw)$ as
\begin{alignat}{2}&
\|\nabla_{\bbs} g(\bbs,\bbt,\bbw)-\nabla_{\bbs} g(\hat{\bbs},\hat{\bbt},\hat{\bbw})\|_2^2
&\nonumber\\&
~\quad\leq 
\lambda_1^2\|\bbPsi^\top\bbPsi\|_F^2\|\bbs-\hat{\bbs}\|_2^2 
 + \left(1/\epsilon^4 + \lambda_1^2\|\bbPsi^\top\|_F^2\right)\|\bbt-\hat{\bbt}\|_2^2
&\nonumber\\&
~\quad\leq 
\lambda_1^2G_s^s\|\bbs-\hat{\bbs}\|_2^2 + \left(1/\epsilon^4 + \lambda_1^2G_s^t\right)\|\bbt-\hat{\bbt}\|_2^2,
&\nonumber
\end{alignat}
where $G_s^s = \|\bbPsi^\top\bbPsi\|_F^2$ and $G_s^t = \|\bbPsi^\top\|_F^2$.
Similarly, for the gradient of $g(\bbs,\bbt,\bbw)$ with respect to $\bbw$, we have
\begin{alignat}{2}&
\|\nabla_{\bbw} g(\bbs,\bbt,\bbw)-\nabla_{\bbw} g(\hat{\bbs},\hat{\bbt},\hat{\bbw})\|_2^2
&\nonumber\\&
~~\quad\leq 
\lambda_2^2\|\bbSigma^\top\bbSigma\|_F^2\|\bbw-\hat{\bbw}\|_2^2 
 + \lambda_2^2\|\bbSigma^\top\|_F^2\|\bbt-\hat{\bbt}\|_2^2
&\nonumber\\&
~~\quad\leq 
\lambda_2^2G_w^w\|\bbw-\hat{\bbw}\|_2^2 + \lambda_2^2G_w^t\|\bbt-\hat{\bbt}\|_2^2,
&\nonumber
\end{alignat}
where $G_w^w = \|\bbSigma^\top\bbSigma\|_F^2$ and $G_w^t = \|\bbSigma^\top\|_F^2$.

We bound the gradient difference with respect to $\bbt$ with fixed $\bbs$ and $\bbt$ as 
\begin{alignat}{3}&
\|\nabla_{\bbt} g(\bbs,\bbt,\bbw)-\nabla_{\bbt} g(\bbs,\bbt,\hat{\bbw})\|_2^2
&~\leq~& \lambda_2^2\|\bbSigma\|_F^2\|\bbw-\hat{\bbw}\|_2^2
&\nonumber\\&
&~\leq~& \lambda_2^2G_w^t\|\bbw-\hat{\bbw}\|_2^2,
&\nonumber
\end{alignat}
with $G_w^t$ defined previously.

The bounds for part (b) are found similarly, where
\begin{alignat}{3}&
\|\nabla_{\bbs} h(\bbs,\bbw)-\nabla_{\bbs} h(\hat{\bbs},\hat{\bbw})\|_2^2
&~\leq~& \alpha^2\|\bbM^\top\bbM\|_F^2\|\bbs-\hat{\bbs}\|_2^2
&\nonumber\\&
&~\leq~& \alpha^2H_s\|\bbs-\hat{\bbs}\|_2^2,
&\nonumber
\end{alignat}
\begin{alignat}{3}&
\|\nabla_{\bbw} h(\bbs,\bbw)-\nabla_{\bbw} h(\hat{\bbs},\hat{\bbw})\|_2^2
&~\leq~& \beta^2\|\bbPhi^\top\bbPhi\|_F^2\|\bbw-\hat{\bbw}\|_2^2
&\nonumber\\&
&~\leq~& \beta^2H_w\|\bbw-\hat{\bbw}\|_2^2,
&\nonumber
\end{alignat}
where $H_s = \|\bbM^\top\bbM\|_F^2$ and $H_w = \|\bbPhi^\top\bbPhi\|_F^2$.
\hfill$\blacksquare$

\medskip

\noindent{\bf Proof of Lemma 3.}
Since $\bbs^j$ and $\bbw^j$ are optimal with respect to (15a) and (15d) at iteration $j$, we have that $0 = \nabla_{\bbs}h(\bbs^j,\bbw^{j-1}) + \nabla_{\bbs}g(\bbs^j,\bbt^{j-1},\bbw^{j-1}) + \rho_1\bbu_1^{j-1} + \rho_1(\bbs^j-\bbp^{j-1})$ and $0 = \nabla_{\bbw}h(\bbs^j,\bbw^j) + \nabla_{\bbw}g(\bbs^j,\bbt^j,\bbw^j) + \rho_2\bbu_2^{j-1} + \rho_2(\bbw^j-\bbv^{j-1})$.
We also have that $\bbu_1^j = \bbu_1^{j-1} + \bbs^j-\bbp^j$ and $\bbu_2^j = \bbu_2^{j-1} + \bbw^j-\bbv^j$. Then, parts (a) and (b) follow.

Next, by Lemma 2 and parts (a) and (b), we can show that (c) holds by
\begin{alignat}{2}&
\|\bbu_1^j-\bbu_1^{j+1}\|_2^2 \leq \frac{1}{\rho_1^2}\|\nabla_{\bbs}h(\bbs^{j+1},\bbw^j)-\nabla_{\bbs}h(\bbs^j,\bbw^{j-1})\|_2^2
&\nonumber\\&
\quad\qquad + \frac{1}{\rho_1^2}\|\nabla_{\bbs}g(\bbs^{j+1},\bbt^j,\bbw^j)-\nabla_{\bbs}g(\bbs^j,\bbt^{j-1},\bbw^{j-1})\|_2^2
&\nonumber\\&
\quad\qquad + \|(\bbp^{j-1}-\bbp^j) - (\bbp^j-\bbp^{j+1})\|_2^2
&\nonumber\\&
\quad\leq \frac{\lambda_1^2G_s^s+\alpha^2H_s}{\rho_1^2}\|\bbs^j-\bbs^{j+1}\|_2^2
&\nonumber\\&
\quad\qquad + \frac{1/\epsilon^4+\lambda_1^2G_s^t}{\rho_1^2}\|\bbt^{j-1}-\bbt^j\|_2^2
&\nonumber\\&
\quad\qquad + \|\bbp^{j-1}-\bbp^j\|_2^2 + \|\bbp^j-\bbp^{j+1}\|_2^2
&\nonumber
\end{alignat}
and (d) holds by
\begin{alignat}{2}&
\|\bbu_2^j-\bbu_2^{j+1}\|_2^2 \leq \frac{1}{\rho_2^2}\|\nabla_{\bbw}h(\bbs^{j+1},\bbw^{j+1})-\nabla_{\bbw}h(\bbs^j,\bbw^j)\|_2^2
&\nonumber\\&
\quad\qquad + \frac{1}{\rho_2^2}\|\nabla_{\bbw}g(\bbs^{j+1},\bbt^{j+1},\bbw^{j+1})-\nabla_{\bbw}g(\bbs^j,\bbt^j,\bbw^j)\|_2^2
&\nonumber\\&
\quad\qquad + \|(\bbv^{j-1}-\bbv^j) - (\bbv^j-\bbv^{j+1})\|_2^2
&\nonumber\\&
\quad \leq \frac{\lambda_2^2G_w^w+\beta^2H_w}{\rho_2^2}\|\bbw^j-\bbw^{j+1}\|_2^2 + \frac{\lambda_2^2G_w^t}{\rho_2^2}\|\bbt^j-\bbt^{j+1}\|_2^2
&\nonumber\\&
\quad\qquad + \|\bbv^{j-1}-\bbv^j\|_2^2 + \|\bbv^j-\bbv^{j+1}\|_2^2,
&\nonumber
\end{alignat}
verifying the desired upper bounds.
\hfill$\blacksquare$

\medskip

\noindent{\bf Proof of Property 1.}
We first show that $\mathfrak{L}_{\bbrho}$ is monotonically decreasing for all $j\in\mathbb{N}$, which will lead to the sequence $\{\bbs^j,\bbp^j,\bbt^j,\bbw^j,\bbv^j,\bbu_1^j,\bbu_2^j\}$ being bounded. 
By the optimality of each primal update  (15a)-(15e), we first have that
%
\begin{alignat}{2}&
\mathfrak{L}_{\bbrho}(\bbs^j,\bbp^j,\bbt^j,\bbw^j,\bbv^j,\bbu_1^j,\bbu_2^j)
&\nonumber\\&
\qquad\qquad \geq \mathfrak{L}_{\bbrho}(\bbs^{j_1},\bbp^{j_2},\bbt^{j_3},\bbw^{j_4},\bbv^{j_5},\bbu_1^j,\bbu_2^j)
&\nonumber
\end{alignat}
for $j_1\geq j_2\geq j_3\geq j_4\geq j_5$ and $j_i\in\{j,j+1\}$ for all $i$.

Furthermore, it is clear that $\mathfrak{L}_{\bbrho}(\bbs,\bbp,\bbt,\bbw,\bbv,\bbu_1,\bbu_2)$ is strongly convex with respect to $\bbs$ and $\bbw$, and $\mathfrak{L}_{\bbrho}(\bbs,\bbp,\bbt,\bbw,\bbv,\bbu_1,\bbu_2)-f(\bbp,\bbv)$ is strongly convex with respect to $\bbp$ and $\bbv$.
Specifically, we have that 
\begin{alignat}{2}&
\mathfrak{L}_{\bbrho}(\bbs^j,\bbp^j,\bbt^j,\bbw^j,\bbv^j,\bbu_1^j,\bbu_2^j)
&\nonumber\\&
\qquad -\mathfrak{L}_{\bbrho}(\bbs^{j+1},\bbp^j,\bbt^j,\bbw^j,\bbv^j,\bbu_1^j,\bbu_2^j)
\geq \frac{\rho_1}{2}\|\bbs^j-\bbs^{j+1}\|_2^2,
\nonumber
\end{alignat}
where the inequality results from the optimality of $\bbs^{j+1}$, that is, $\nabla_{\bbs}\mathfrak{L}_{\rho}(\bbs^{j+1},\bbp^j,\bbt^j,\bbw^j,\bbv^j,\bbu_1^j,\bbu_2^j)=0$. 
Analogously for $\bbw$, we have
\begin{alignat}{2}&
\mathfrak{L}_{\bbrho}(\bbs^{j+1},\bbp^{j+1},\bbt^{j+1},\bbw^j,\bbv^j,\bbu_1^j,\bbu_2^j)
&\nonumber\\&
\qquad\qquad -\mathfrak{L}_{\bbrho}(\bbs^{j+1},\bbp^{j+1},\bbt^{j+1},\bbw^{j+1},\bbv^j,\bbu_1^j,\bbu_2^j)
&\nonumber\\&
\qquad \geq \frac{\rho_2}{2}\|\bbw^j-\bbw^{j+1}\|_2^2.
\nonumber
\end{alignat}

By the optimality of $\bbp^{j+1}$, we have that
$0\in\partial_{\bbp}f(\bbp^{j+1},\bbv^{j}) - \rho_1\bbu_1^{j+1}$, so $\rho_1\bbu_1^{j+1}\in\partial_{\bbp}f(\bbp^{j+1},\bbv^{j})$.
By the definition of the subgradient, we have that $\langle\rho_1\bbu_1^{j+1},\bbp^{j+1}\rangle\geq\langle\rho_1\bbu_1^{j+1},\bbp\rangle$ for any $\bbp\in\{0,1\}^{L_K}$.
%
It then follows that $\rho_1\langle\bbu_1^{j+1},\bbp^{j+1}-\bbp^{i}\rangle\geq0$ for all $i\in\mathbb{N}$.
By strong convexity of $\mathfrak{L}_{\bbrho}(\bbs,\bbp,\bbt,\bbw,\bbv,\bbu_1,\bbu_2)-f(\bbp,\bbv)$ with respect to $\bbp$,
\begin{alignat}{3}&
\rho_1\langle \bbu_1^{j+1},\bbp^{j+1}-\bbp^j \rangle 
&~\geq~&
\rho_1\langle \bbu_1^{j+1}-\bbu_1^j,\bbp^{j+1}-\bbp^j \rangle 
&\nonumber\\&
&~\geq~&
\rho_1\|\bbp^j-\bbp^{j+1}\|_2^2.
&\nonumber
\end{alignat}
By an analogous process for $\bbv$, we can also show that $\rho_2\langle \bbu_2^{j+1},\bbv^{j+1}-\bbv^j \rangle \geq \rho_2\|\bbv^j-\bbv^{j+1}\|_2^2$.

Finally, since $\nabla_{\bbt}g(\bbs^{j+1},\bbt^{j+1},\bbw^j)=0$ by optimality of $\bbt^{j+1}$, we have
\begin{alignat}{3}&
\mathfrak{L}_{\bbrho}(\bbs^{j+1},\bbp^{j+1},\bbt^j,\bbw^j,\bbv^j,\bbu_1^j,\bbu_2^j)
&\nonumber\\&
\qquad\qquad -\mathfrak{L}_{\bbrho}(\bbs^{j+1},\bbp^{j+1},\bbt^{j+1},\bbw^j,\bbv^j,\bbu_1^j,\bbu_2^j)
&\nonumber\\&
\qquad= g(\bbs^{j+1},\bbt^{j},\bbw^{j})-g(\bbs^{j+1},\bbt^{j+1},\bbw^{j})
&\nonumber\\&
\qquad= \bbGamma(\bbs^{j+1},\bbt^j) - \bbGamma(\bbs^{j+1},\bbt^{j+1})
&\nonumber\\&
\qquad\qquad - \langle \nabla_{\bbt}\bbGamma(\bbs^{j+1},\bbt^{j+1}),\bbt^j-\bbt^{j+1} \rangle
&\nonumber\\&
\qquad\qquad + \langle \nabla_{\bbt}g(\bbs^{j+1},\bbt^{j+1},\bbw^{j}),\bbt^j-\bbt^{j+1} \rangle
&\nonumber\\&
\qquad\qquad + \frac{\lambda_1+\lambda_2}{2}\|\bbt^j-\bbt^{j+1}\|_2^2
&\nonumber\\&
\qquad= \bbGamma(\bbs^{j+1},\bbt^j) - \bbGamma(\bbs^{j+1},\bbt^{j+1})
&\nonumber\\&
\qquad\qquad - \langle \nabla_{\bbt}\bbGamma(\bbs^{j+1},\bbt^{j+1}),\bbt^j-\bbt^{j+1} \rangle
&\nonumber\\&
\qquad\qquad + \frac{\lambda_1+\lambda_2}{2}\|\bbt^j-\bbt^{j+1}\|_2^2
&\nonumber\\&
\qquad\geq \left(\frac{\lambda_1+\lambda_2}{2} - \frac{1}{2\epsilon^4}\right)\|\bbt^j-\bbt^{j+1}\|_2^2,
&\nonumber
\end{alignat}
where the inequality results since $\bbGamma(\bbs,\bbt)$ is $1/\epsilon^4$-smooth.

Combining the previously derived lower bounds and Lemma 3 for the full update of $\mathfrak{L}_{\bbrho}$, we have
\begin{alignat}{2}&
\mathfrak{L}_{\bbrho}(\bbs^j,\bbp^j,\bbt^j,\bbw^j,\bbv^j,\bbu_1^j,\bbu_2^j)
&\nonumber\\&
\qquad\qquad -\mathfrak{L}_{\bbrho}(\bbs^{j+1},\bbp^{j+1},\bbt^{j+1},\bbw^{j+1},\bbv^{j+1},\bbu_1^{j+1},\bbu_2^{j+1})
&\nonumber\\&
\qquad\geq \left(\frac{\rho_1}{2}-\frac{\lambda_1^2G_s^s+\alpha^2H_s}{\rho_1}\right)\|\bbs^j-\bbs^{j+1}\|_2^2
&\nonumber\\&
\qquad\qquad+ \left(\frac{\rho_2}{2}-\frac{\lambda_2^2G_w^w+\beta^2H_w}{\rho_2}\right)\|\bbw^j-\bbw^{j+1}\|_2^2
&\nonumber\\&
\qquad\qquad+ \left(\frac{\lambda_1+\lambda_2}{2}-\frac{1}{2\epsilon^4}-\frac{\lambda_2^2G_w^t}{\rho_2}\right)\|\bbt^j-\bbt^{j+1}\|_2^2
&\nonumber\\&
\qquad\qquad - \frac{\epsilon^{-4}+\lambda_1^2G_s^t}{\rho_1}\|\bbt^{j-1}-\bbt^j\|_2^2
&\nonumber\\&
\qquad\qquad+ \frac{\rho_1}{2}\|\bbp^j-\bbp^{j+1}\|_2^2 
- \frac{\rho_1}{2}\|\bbp^{j-1}-\bbp^j\|_2^2
&\nonumber\\&
\qquad\qquad+ \frac{\rho_2}{2}\|\bbv^j-\bbv^{j+1}\|_2^2 
- \frac{\rho_2}{2}\|\bbv^{j-1}-\bbv^j\|_2^2.
&\nonumber
\end{alignat}
Summing over all iterates up to a given $j\in\mathbb{N}$, we have the following difference
\begin{alignat}{2}&
\mathfrak{L}_{\bbrho}(\bbs^{0},\bbp^{0},\bbt^{0},\bbw^{0},\bbv^{0},\bbu_1^{0},\bbu_2^{0})
&\nonumber\\&
\quad\quad-\mathfrak{L}_{\bbrho}(\bbs^j,\bbp^j,\bbt^j,\bbw^j,\bbv^j,\bbu_1^j,\bbu_2^j)
&\nonumber\\&
\quad\geq \sum_{i=0}^j \left(\frac{\rho_1}{2}-\frac{\lambda_1^2G_s^s+\alpha^2H_s}{\rho_1}\right)\|\bbs^{i}-\bbs^{i+1}\|_2^2
&\nonumber\\&
\quad\quad + \sum_{i=0}^j \left(\frac{\rho_2}{2}-\frac{\lambda_2^2G_w^w+\beta^2H_w}{\rho_2}\right)\|\bbw^{i}-\bbw^{i+1}\|_2^2
&\nonumber\\&
\quad\quad + \sum_{i=0}^j \bigg(\frac{\lambda_1+\lambda_2}{2}-\frac{1}{2\epsilon^4} \bigg.
&\nonumber\\&
\quad\qquad\qquad \bigg. -\frac{\epsilon^{-4}+\lambda_1^2G_s^t}{\rho_1}-\frac{\lambda_2^2G_w^t}{\rho_2}\bigg)\|\bbt^i-\bbt^{i+1}\|_2^2
&\nonumber\\&
\quad\quad + \sum_{i=0}^j \frac{\rho_1}{2}\|\bbp^{i}-\bbp^{i+1}\|_2^2
 + \sum_{i=0}^j \frac{\rho_2}{2}\|\bbv^{i}-\bbv^{i+1}\|_2^2,
&\label{E:suff_desc}
\end{alignat}
where $\rho_1$ and $\rho_2$ are chosen to be large enough so that coefficients of $\|\bbs^{i}-\bbs^{i+1}\|_2^2$ and $\|\bbw^{i}-\bbw^{i+1}\|_2^2$ are nonnegative, and $\lambda_1$ and $\lambda_2$ are large enough so that the coefficient of $\|\bbt^i-\bbt^{i+1}\|_2^2$ is nonnegative.
Then, we can guarantee that $\mathfrak{L}_{\bbrho}(\bbs^j,\bbp^j,\bbt^j,\bbw^j,\bbv^j,\bbu_1^j,\bbu_2^j)$ is upper bounded by $\mathfrak{L}_{\bbrho}(\bbs^{0},\bbp^{0},\bbt^{0},\bbw^{0},\bbv^{0},\bbu_1^{0},\bbu_2^{0})$.
Thus, $\phi(\bbs^j,\bbp^j,\bbt^j,\bbw^j,\bbv^j)$, $\|\bbs^j-\bbp^j\|_2^2$, and $\|\bbw^j-\bbv^j\|_2^2$ are also upper bounded.
It then follows that $\{\bbs^j,\bbp^j,\bbt^j,\bbw^j,\bbv^j\}$ is bounded due to Lemma 1.
By Lemma 3, we conclude that $\{\bbu_1^j,\bbu_2^j\}$ is also bounded. 

Finally, to show that $\mathfrak{L}_{\bbrho}(\bbs^j,\bbp^j,\bbt^j,\bbw^j,\bbv^j,\bbu_1^j,\bbu_2^j)$ is lower bounded for all $j\in\mathbb{N}$, let us introduce $\bbw'=\bbv^j$.
We have that $g(\bbs^j,\bbt^j,\bbw') + h(\bbs^j,\bbw') + f(\bbp^j,\bbv^j)$ is greater than $\min_{\bbw} g(\bbs^j,\bbt^j,\bbw) + h(\bbs^j,\bbw) + f(\bbp^j,\bbw)$, which is lower bounded by Lemma 1. 
We then have that 
\begin{alignat}{2}&
\mathfrak{L}_{\bbrho}(\bbs^j,\bbp^j,\bbt^j,\bbw^j,\bbv^j,\bbu_1^j,\bbu_2^j)
= g(\bbs^j,\bbt^j,\bbw^j) 
&\nonumber\\&
\qquad\qquad + h(\bbs^j,\bbw^j) + f(\bbp^j,\bbv^j)
&\nonumber\\&
\qquad\qquad + \rho_1 \langle \bbu_1^j, \bbs^j-\bbp^j \rangle + \frac{\rho_1}{2}\|\bbs^j-\bbp^j\|_2^2
&\nonumber\\&
\qquad\qquad + \rho_2 \langle \bbu_2^j, \bbw^j-\bbv^j \rangle + \frac{\rho_2}{2}\|\bbw^j-\bbv^j\|_2^2
&\nonumber\\&
\qquad \geq g(\bbs^j,\bbt^j,\bbw') + h(\bbs^j,\bbw') + f(\bbp^j,\bbv^j)
&\nonumber\\&
\qquad\qquad + \frac{\rho_2-\lambda_2^2G_w^w-\beta^2H_w}{2}\|\bbw^j-\bbw'\|_2^2
&\nonumber\\&
\qquad\qquad + \rho_1 \langle \bbu_1^j, \bbs^j-\bbp^j \rangle + \frac{\rho_1}{2}\|\bbs^j-\bbp^j\|_2^2
&\nonumber\\&
\qquad\qquad + \rho_2\langle\bbu_2^j-\bbu_2^{j-1},\bbv^{j-1}-\bbv^j\rangle > -\infty,
&\nonumber
\end{alignat}
where the inner product terms are lower bounded since $\{\bbs^j,\bbp^j,\bbv^j,\bbu_1^j,\bbu_2^j\}$ is bounded for all $j\in\mathbb{N}$.
Thus, $\mathfrak{L}_{\bbrho}(\bbs^j,\bbp^j,\bbt^j,\bbw^j,\bbv^j,\bbu_1^j,\bbu_2^j)$ is lower bounded. 
\hfill$\blacksquare$

\medskip

\noindent{\bf Proof of Property 2.}
The upper bound in Property 2 follows directly from the bound \eqref{E:suff_desc} in the previous proof. 
\hfill$\blacksquare$

\medskip

\noindent{\bf Proof of Property 3.}
Since the objective consists of terms containing $(\bbs^j,\bbp^j,\bbt^j,\bbw^j,\bbv^j,\bbu_1^j,\bbu_2^j)$, we need only show that each block of $\partial\mathfrak{L}_{\bbrho}$ can be bounded with respect to a controllable constant related to $\bbrho$ or $\bblambda$.

First, we consider the gradient with respect to $\bbs$,
\begin{alignat}{2}&
\nabla_{\bbs}\mathfrak{L}_{\bbrho}(\bbs^{j+1},\bbp^{j+1},\bbt^{j+1},\bbw^{j+1},\bbv^{j+1},\bbu_1^{j+1},\bbu_2^{j+1})
&\nonumber\\&
\qquad= \nabla_{\bbs}h(\bbs^{j+1},\bbw^{j+1}) 
+ \nabla_{\bbs}g(\bbs^{j+1},\bbt^{j+1},\bbw^{j+1}) 
&\nonumber\\&
\qquad\qquad+ \rho_1(\bbs^{j+1}-\bbp^{j+1}+\bbu_1^{j+1})
&\nonumber\\&
\qquad= \rho_1(\bbu_1^{j+1}-\bbu_1^j) + \rho_1(\bbp^j-\bbp^{j+1})
&\nonumber\\&
\qquad\qquad+ \nabla_{\bbs}h(\bbs^{j+1},\bbw^{j+1}) 
+ \nabla_{\bbs}g(\bbs^{j+1},\bbt^{j+1},\bbw^{j+1})
&\nonumber\\&
\qquad\qquad- \nabla_{\bbs}h(\bbs^{j+1},\bbw^j) 
- \nabla_{\bbs}g(\bbs^{j+1},\bbt^j,\bbw^j),
&\nonumber
\end{alignat}
whose magnitude can be bounded as
\begin{alignat}{2}&
\|\nabla_{\bbs}\mathfrak{L}_{\bbrho}(\bbs^{j+1},\bbp^{j+1},\bbt^{j+1},\bbw^{j+1},\bbv^{j+1},\bbu_1^{j+1},\bbu_2^{j+1})\|_2^2
&\nonumber\\&
\quad\leq \rho_1^2\|\bbu_1^j-\bbu_1^{j+1}\|_2^2 + \rho_1^2\|\bbp^j-\bbp^{j+1}\|_2^2
&\nonumber\\&
\quad\qquad
+ \|\nabla_{\bbs}h(\bbs^{j+1},\bbw^{j+1}) 
- \nabla_{\bbs}h(\bbs^{j+1},\bbw^j) \|_2^2
&\nonumber\\&
\quad\qquad
+ \|\nabla_{\bbs}g(\bbs^{j+1},\bbt^{j+1},\bbw^{j+1})
- \nabla_{\bbs}g(\bbs^{j+1},\bbt^j,\bbw^j)\|_2^2,
&\nonumber
\end{alignat}
and, by Lemmas 2 and 3, this results in the bound
\begin{alignat}{2}&
\|\nabla_{\bbs}\mathfrak{L}_{\bbrho}(\bbs^{j+1},\bbp^{j+1},\bbt^{j+1},\bbw^{j+1},\bbv^{j+1},\bbu_1^{j+1},\bbu_2^{j+1})\|_2^2
&\nonumber\\&
\quad\leq (\lambda_1^2G_s^s+\alpha^2H_s)\|\bbs^j-\bbs^{j+1}\|_2^2
&\nonumber\\&
\quad\qquad + (1/\epsilon^4 + \lambda_1^2G_s^t)\left(\|\bbt^j-\bbt^{j+1}\|_2^2
 + \|\bbt^{j-1}-\bbt^j\|_2^2\right)
&\nonumber\\&
\quad\qquad + 2\rho_1^2\|\bbp^j-\bbp^{j+1}\|_2^2
 + \rho_1^2\|\bbp^{j-1}-\bbp^j\|_2^2.
&\nonumber
\end{alignat}
By an analogous derivation, we can obtain the following bound with respect to $\bbw$ as
\begin{alignat}{2}&
\|\nabla_{\bbw}\mathfrak{L}_{\bbrho}(\bbs^{j+1},\bbp^{j+1},\bbt^{j+1},\bbw^{j+1},\bbv^{j+1},\bbu_1^{j+1},\bbu_2^{j+1})\|_2^2
&\nonumber\\&
\quad\leq (\lambda_2^2G_w^w+\beta^2H_w)\|\bbw^j-\bbw^{j+1}\|_2^2
 + \lambda_2^2G_w^t\|\bbt^j-\bbt^{j+1}\|_2^2
&\nonumber\\&
\quad\qquad + 2\rho_2^2\|\bbv^j-\bbv^{j+1}\|_2^2
 + \rho_2^2\|\bbv^{j-1}-\bbv^j\|_2^2.
&\nonumber
\end{alignat}

We recall that $\nabla_{\bbt}g(\bbs^{j+1},\bbt^{j+1},\bbw^j)=0$, so we have
\begin{alignat}{2}&
\nabla_{\bbt}\mathfrak{L}_{\bbrho}(\bbs^{j+1},\bbp^{j+1},\bbt^{j+1},\bbw^{j+1},\bbv^{j+1},\bbu_1^{j+1},\bbu_2^{j+1})
&\nonumber\\&
\qquad= \nabla_{\bbt}g(\bbs^{j+1},\bbt^{j+1},\bbw^{j+1})
&\nonumber\\&
\qquad= \nabla_{\bbt}g(\bbs^{j+1},\bbt^{j+1},\bbw^{j+1})-\nabla_{\bbt}g(\bbs^{j+1},\bbt^{j+1},\bbw^j),
&\nonumber
\end{alignat}
and thus with Lemma 2 we can write
\begin{alignat}{2}&
\|\nabla_{\bbt}\mathfrak{L}_{\bbrho}(\bbs^{j+1},\bbp^{j+1},\bbt^{j+1},\bbw^{j+1},\bbv^{j+1},\bbu_1^{j+1},\bbu_2^{j+1})\|_2^2
&\nonumber\\&
\qquad= \|\nabla_{\bbt}g(\bbs^{j+1},\bbt^{j+1},\bbw^{j+1})-\nabla_{\bbt}g(\bbs^{j+1},\bbt^{j+1},\bbw^j)\|_2^2,
&\nonumber\\&
\qquad\leq \lambda_2^2G_w^t\|\bbw^j-\bbw^{j+1}\|_2^2.
&\nonumber
\end{alignat}
%

\begin{figure*}[!ht]
\centering
	\begin{minipage}[c]{.32\textwidth}
		\includegraphics[width=\textwidth]{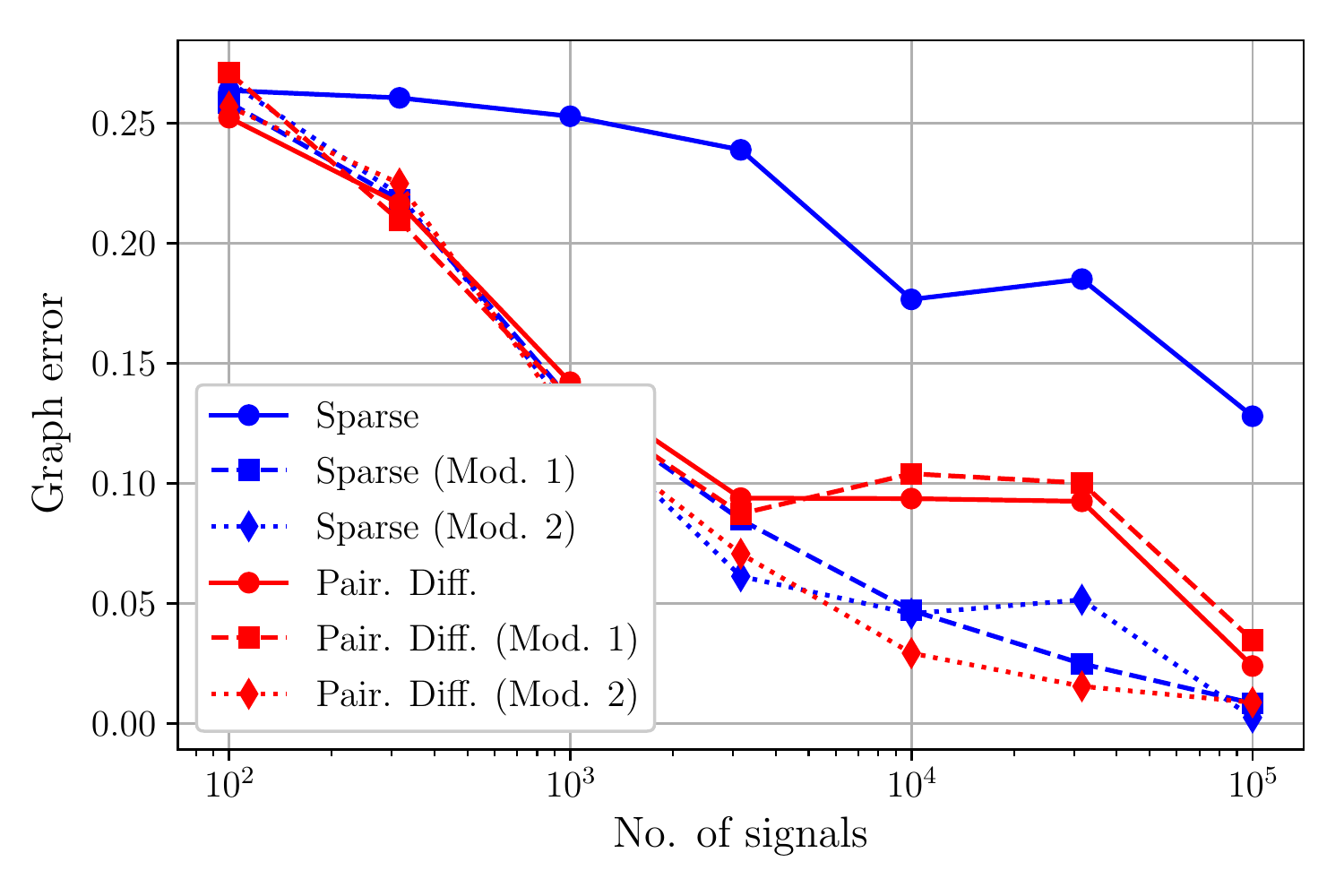}
		\centering{\small (a)}
	\end{minipage}
	\begin{minipage}[c]{.32\textwidth}
		\includegraphics[width=\textwidth]{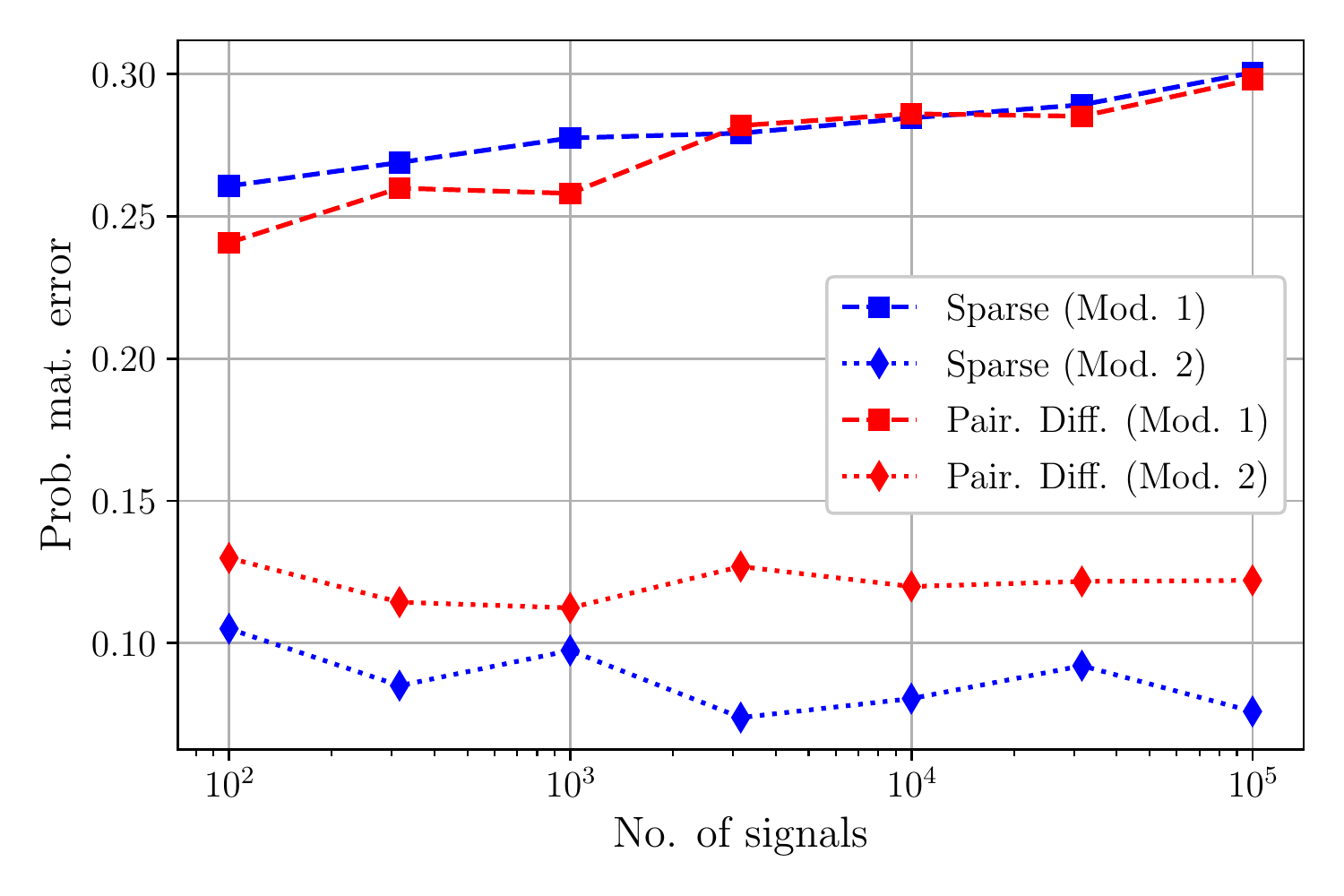}
		\centering{\small (b)}
	\end{minipage}
	\begin{minipage}[c]{.32\textwidth}
		\includegraphics[width=\textwidth]{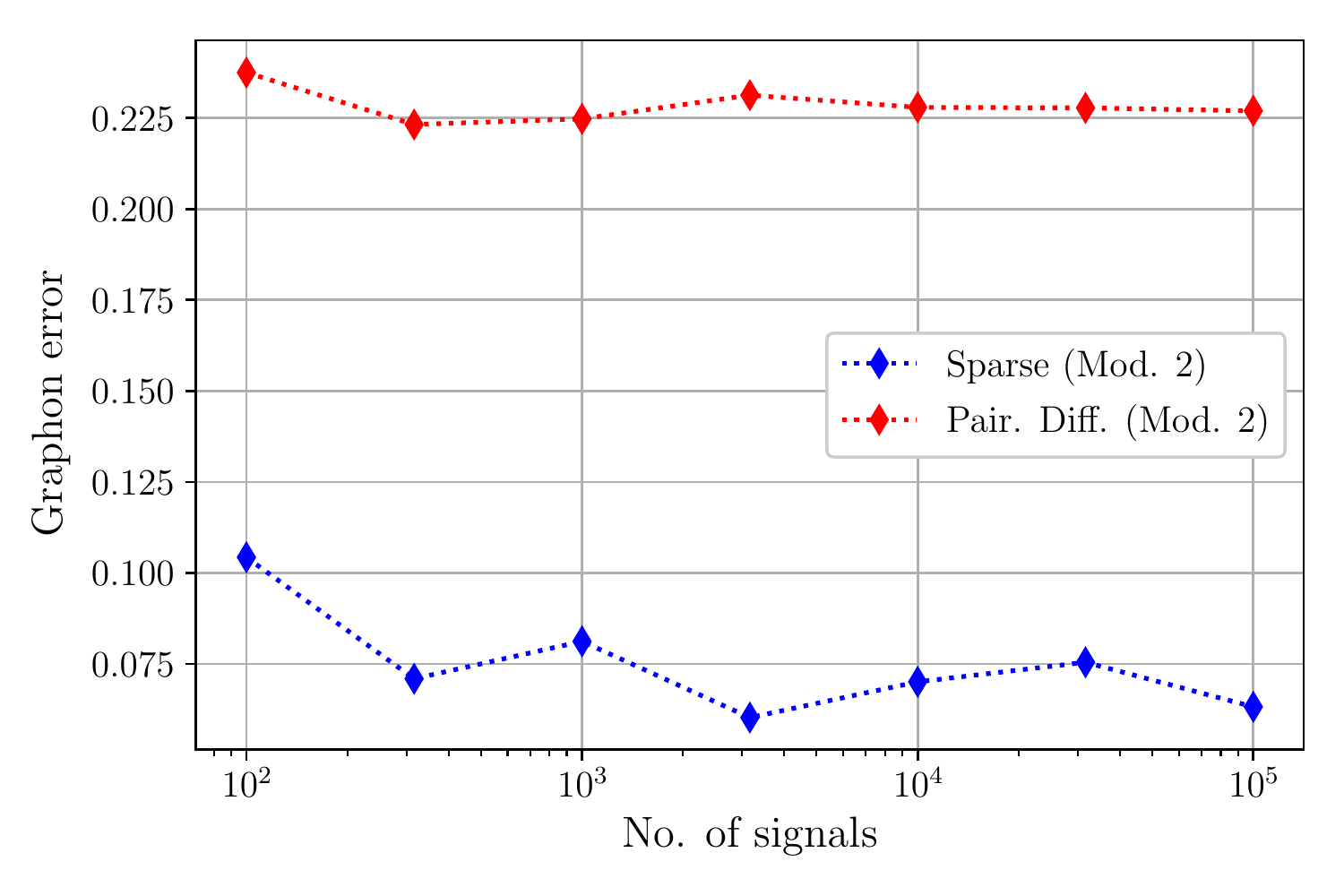}
		\centering{\small (c)}
	\end{minipage}
\caption{ \small{
Performance analysis in synthetic networks sampled from {\it different latent point sets} in the same graphon. 
(a)~Recovery error for $K=3$ graphs of the same size $N=30$ as a function of the number of observed signals. 
(b)~Recover error for $K=3$ probability matrices as subsets of the generating graphon at the latent sample points.
(c)~Recover error for the graphon. 
}}
\label{fig_allerrors}
\end{figure*}

The bounds for $\bbu_1$ and $\bbu_2$ are obtained similarly, where
\begin{alignat}{2}&
\nabla_{\bbu_1}\mathfrak{L}_{\bbrho}(\bbs^{j+1},\bbp^{j+1},\bbt^{j+1},\bbw^{j+1},\bbv^{j+1},\bbu_1^{j+1},\bbu_2^{j+1})
&\nonumber\\&
\qquad= \rho_1(\bbs^{j+1}-\bbp^{j+1})
= \rho_1(\bbu_1^{j+1}-\bbu_1^j)
&\nonumber
\end{alignat}
which is bounded via Lemma 3 as
\begin{alignat}{2}&
\|\nabla_{\bbu_1}\mathfrak{L}_{\bbrho}(\bbs^{j+1},\bbp^{j+1},\bbt^{j+1},\bbw^{j+1},\bbv^{j+1},\bbu_1^{j+1},\bbu_2^{j+1})\|_2^2
&\nonumber\\&
\qquad\leq (\lambda_1^2G_s^s+\alpha^2H_s)\|\bbs^j-\bbs^{j+1}\|_2^2
&\nonumber\\&
\qquad\qquad+ (1/\epsilon^4+\lambda_1^2G_s^t)\|\bbt^{j-1}-\bbt^j\|_2^2
&\nonumber\\&
\qquad\qquad+ \rho_1^2\|\bbp^j-\bbp^{j+1}\|_2^2
+ \rho_1^2\|\bbp^{j-1}-\bbp^j\|_2^2,
&\label{E:subdiff_bound_u1}
\end{alignat}
thus, similarly, we have the bound with respect to $\bbu_2$ as 
\begin{alignat}{2}&
\|\nabla_{\bbu_2}\mathfrak{L}_{\bbrho}(\bbs^{j+1},\bbp^{j+1},\bbt^{j+1},\bbw^{j+1},\bbv^{j+1},\bbu_1^{j+1},\bbu_2^{j+1})\|_2^2
&\nonumber\\&
\quad\leq (\lambda_2^2G_w^w+\beta^2H_w)\|\bbw^j-\bbw^{j+1}\|_2^2
+ \lambda_2^2G_w^t\|\bbt^j-\bbt^{j+1}\|_2^2
&\nonumber\\&
\quad\qquad+ \rho_2^2\|\bbv^j-\bbv^{j+1}\|_2^2
+ \rho_2^2\|\bbv^{j-1}-\bbv^j\|_2^2.
&\label{E:subdiff_bound_u2}
\end{alignat}

For $\bbp$, we first observe the subdifferential 
\begin{alignat}{2}&
\partial_{\bbp} \mathfrak{L}_{\bbrho}(\bbs^{j+1},\bbp^{j+1},\bbt^{j+1},\bbw^{j+1},\bbv^{j+1},\bbu_1^{j+1},\bbu_2^{j+1})
&\nonumber\\&
\qquad= \partial_{\bbp}f(\bbp^{j+1},\bbv^{j+1})
-\rho_1(\bbs^{j+1}-\bbp^{j+1}+\bbu_1^{j+1})
&\nonumber\\&
\qquad= \partial_{\bbp}f(\bbp^{j+1},\bbv^{j+1})
-\rho_1(\bbs^{j+1}-\bbp^{j+1}+\bbu_1^j)
&\nonumber\\&
\qquad\qquad+\rho_1(\bbu_1^j-\bbu_1^{j+1}),
&\nonumber
\end{alignat}
%
and by the optimality of the $\bbp$ update, $0\in\partial_{\bbp}f(\bbp^{j+1},\bbv^{j+1})-\rho_1(\bbs^{j+1}-\bbp^{j+1}+\bbu_1^j)$, so we have the subgradient 
\begin{alignat}{2}&
\bbd_{\bbp} = \rho_1(\bbu_1^j-\bbu_1^{j+1}) \in\partial_{\bbp}\mathfrak{L}_{\bbrho},
&\nonumber
\end{alignat}
which does not depend on $\bbp$.
Note that $\|\bbd_{\bbp}\|_2^2$ is upper bounded as in \eqref{E:subdiff_bound_u1}, and similarly $\|\bbd_{\bbv}\|_2^2$ is upper bounded as in \eqref{E:subdiff_bound_u2}.
Thus, we have that the subgradient of $\mathfrak{L}_{\bbrho}$ is bounded with respect to the iterative updates.
\hfill$\blacksquare$

\medskip

\noindent{\bf Proof of Property 4.}
Since the subsequence $\mathfrak{L}_{\bbrho}(\bbs^{j_a},\bbp^{j_a},\bbt^{j_a},\bbw^{j_a},\bbv^{j_a},\bbu_1^{j_a},\bbu_2^{j_a})$ is lower bounded and asymptotically monotonically nonincreasing, the sequence is convergent. 
Then, we have that the limit $\lim_{a\rightarrow\infty}\mathfrak{L}_{\bbrho}(\bbs^{j_a},\bbp^{j_a},\bbt^{j_a},\bbw^{j_a},\bbv^{j_a},\bbu_1^{j_a},\bbu_2^{j_a})\geq\mathfrak{L}_{\bbrho}(\bbs^{*},\bbp^{*},\bbt^{*},\bbw^{*},\bbv^{*},\bbu_1^{*},\bbu_2^{*})$.
The only potentially discontinuous terms in $\mathfrak{L}_{\bbrho}(\bbs^{j_a},\bbp^{j_a},\bbt^{j_a},\bbw^{j_a},\bbv^{j_a},\bbu_1^{j_a},\bbu_2^{j_a})-\mathfrak{L}_{\bbrho}(\bbs^{*},\bbp^{*},\bbt^{*},\bbw^{*},\bbv^{*},\bbu_1^{*},\bbu_2^{*})$ would be $f(\bbp^{j_a},\bbv^{j_a})-f(\bbp^{*},\bbv^{*})$.
However, since $f(\bbp^j,\bbv^j)=0$ for all $j\in\mathbb{N}$ due to the projections in update steps (15b) and (15e), we also have that $f(\bbp^{j_a},\bbv^{j_a})=0$ for all $a\in\mathbb{N}$, so $f(\bbp^{j_a},\bbv^{j_a})-f(\bbp^{*},\bbv^{*})\leq 0$.
Thus, we have that $\lim_{a\rightarrow\infty}\mathfrak{L}_{\bbrho}(\bbs^{j_a},\bbp^{j_a},\bbt^{j_a},\bbw^{j_a},\bbv^{j_a},\bbu_1^{j_a},\bbu_2^{j_a})=\mathfrak{L}_{\bbrho}(\bbs^{*},\bbp^{*},\bbt^{*},\bbw^{*},\bbv^{*},\bbu_1^{*},\bbu_2^{*})$, and thus the subsequence converges to a limit point as $a\rightarrow\infty$.
\hfill$\blacksquare$

\medskip

\begin{figure*}[!ht]
    \centering
        \includegraphics[width=.65\textwidth]{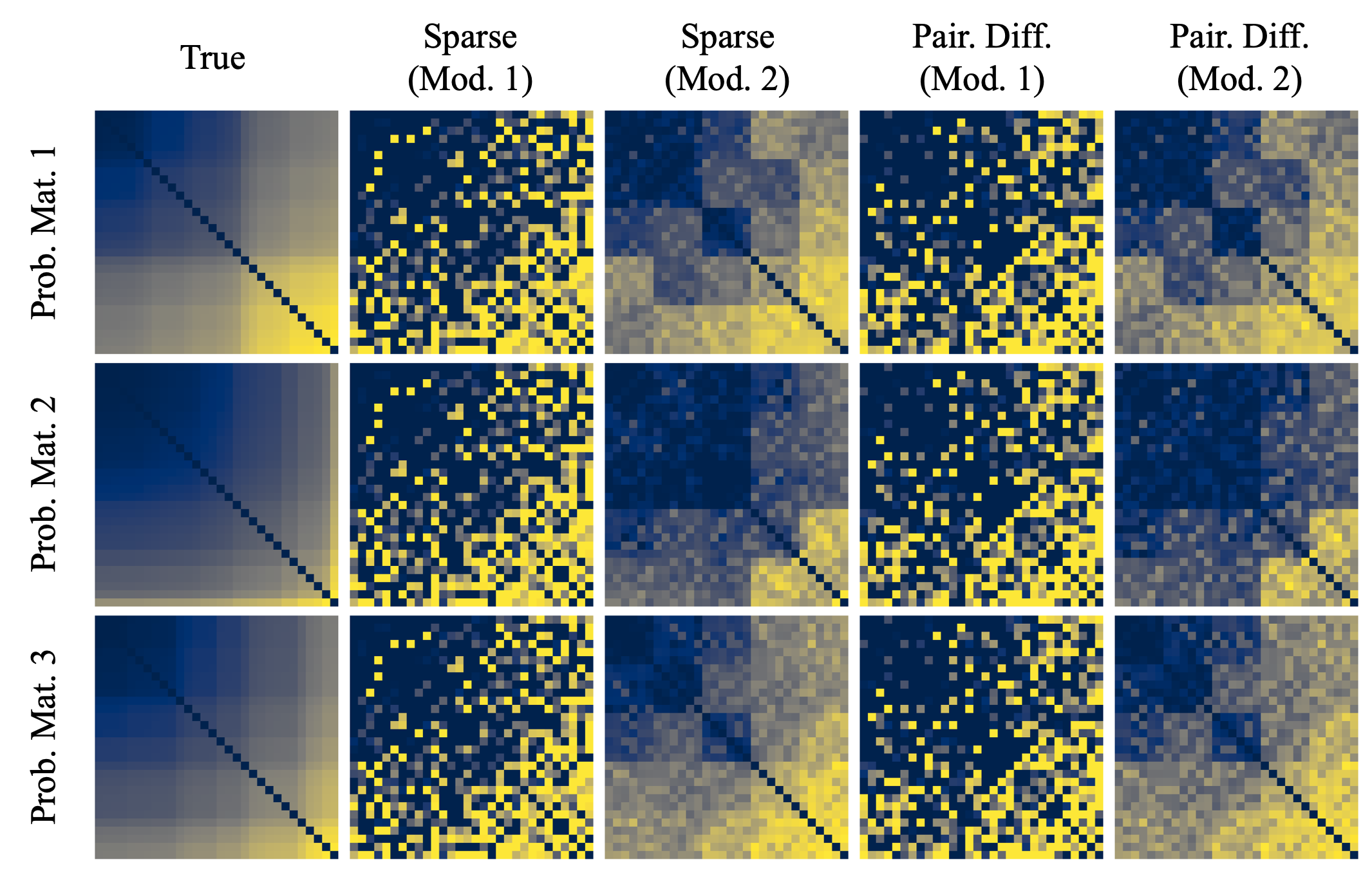}
    \caption{\small{Comparison of estimated probability matrices. The image shows pixel pictures of probability matrices whose entries represent estimated edge probability for each node pair in the networks.}}
    \label{fig_t_figs}
\end{figure*}

\begin{figure*}[!hb]
\centering
	\begin{minipage}[c]{.32\textwidth}
		\includegraphics[width=1.1\textwidth]{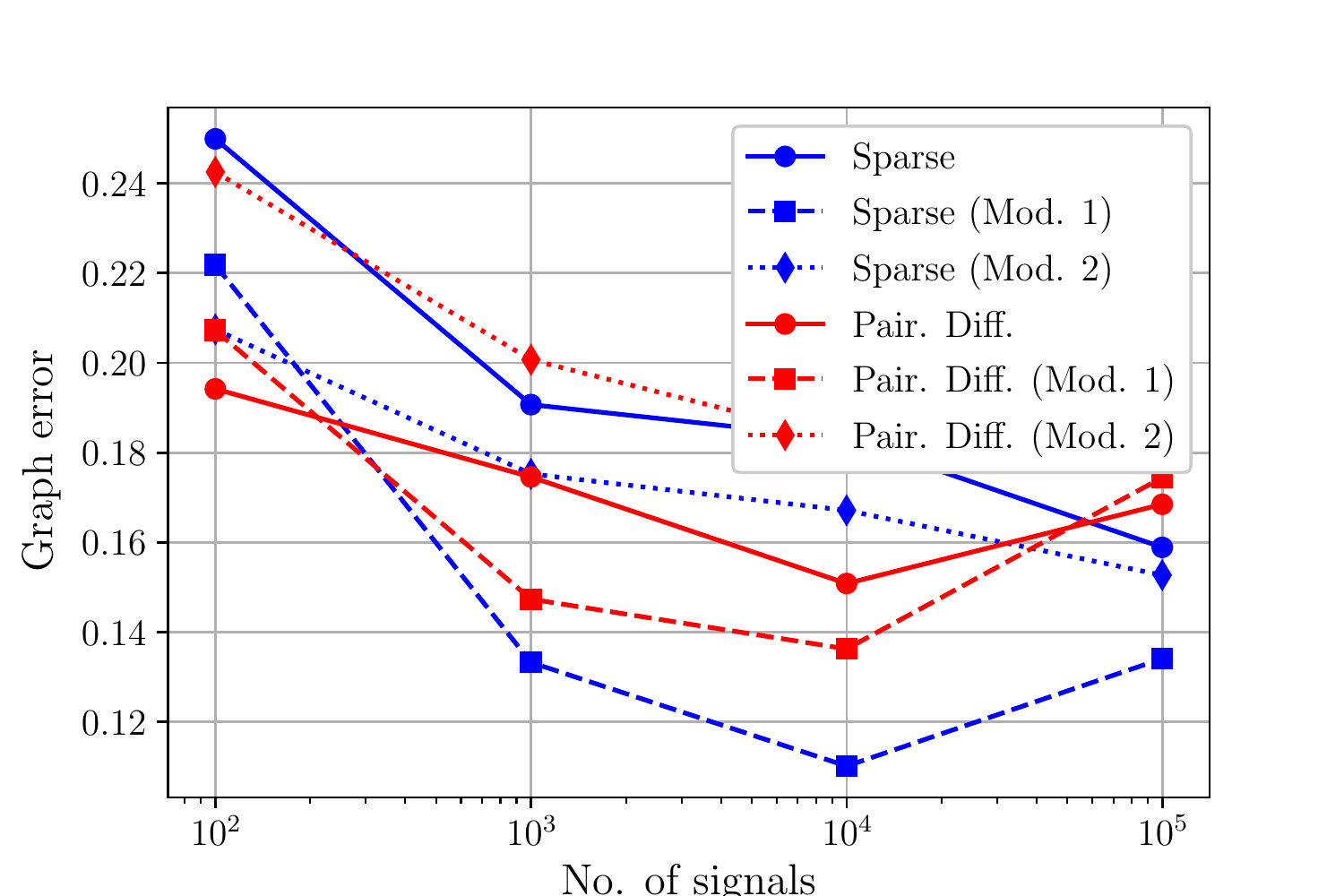}
		\centering{\small (a)}
	\end{minipage}
	\begin{minipage}[c]{.32\textwidth}
		\includegraphics[width=1.1\textwidth]{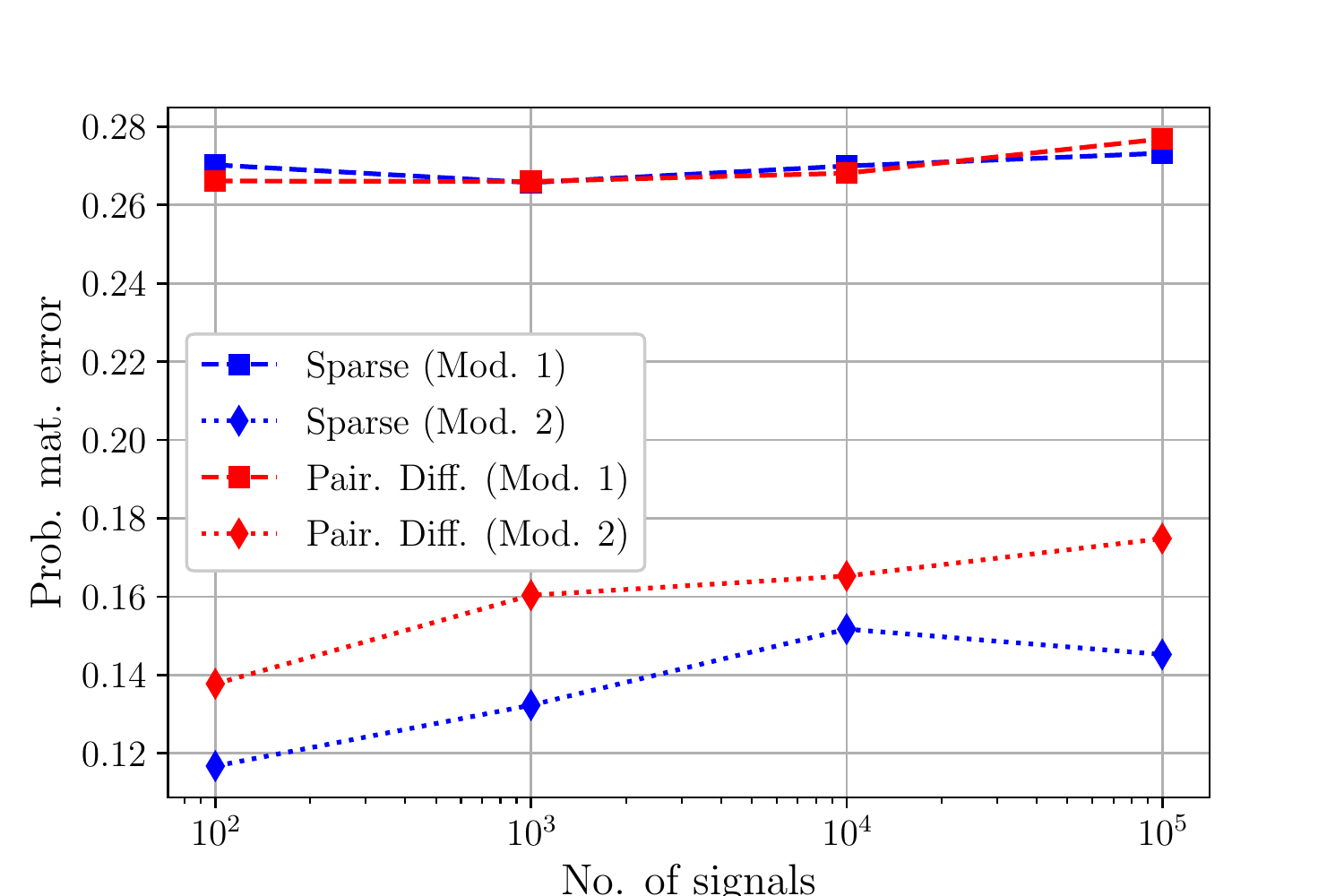}
		\centering{\small (b)}
	\end{minipage}
	\begin{minipage}[c]{.32\textwidth}
		\includegraphics[width=1.1\textwidth]{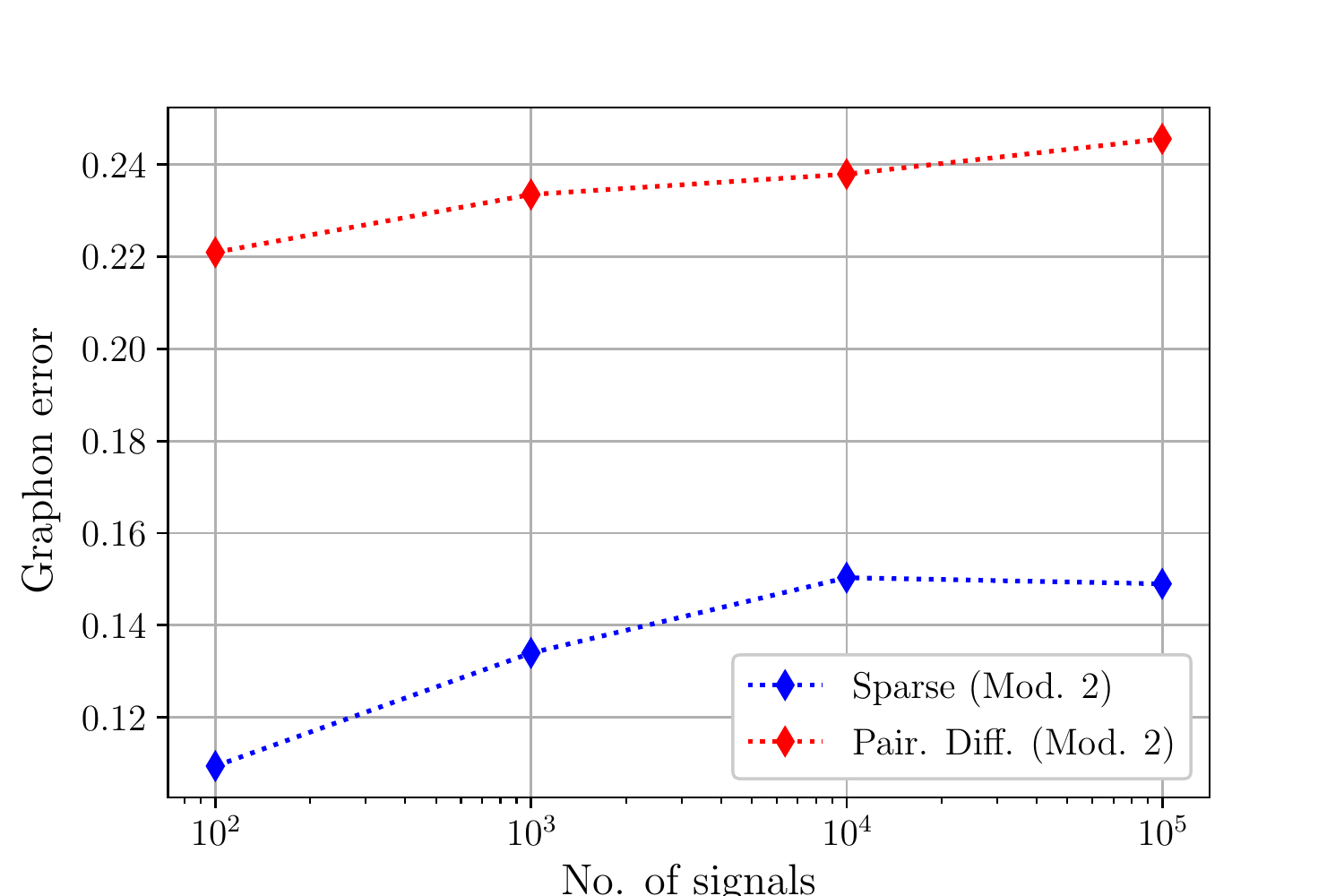}
		\centering{\small (c)}
	\end{minipage}
\caption{ \small{
Performance analysis in \emph{weighted} synthetic networks sampled from the \emph{same latent point sets} in the same graphon.
(a)~Recovery error for $K=3$ graphs as a function of the number of observed signals. 
(b)~Recovery error for $K=3$ probability matrices as subsets of the generating graphon at the latent sample points. 
(c)~Recovery error for the estimated graphon. 
}}
\label{fig_edge_weights}
\end{figure*}

\section{Additional Numerical Experiments}

We include additional results to empirically illustrate the effects of jointly inferring the networks, probability matrices, and graphon.
We also demonstrate the practicality of our proposed method for weighted networks.

\vspace{.1in}
\noindent{\bf Joint estimation of networks, probability matrices, and graphon.}
Observing the empirical results provides us with insight into how each of the three object types\textemdash networks, probability matrices, and graphons\textemdash interact when jointly inferred.
In Fig.~\ref{fig_allerrors}, we consider estimating multiple networks of the \emph{same size} sampled from the same graphon at \emph{different latent points}.
While all proposed augmentations achieve significant improvement in network recovery, jointly estimating the graphon achieves the greatest improvement for both separate and joint network inference in Fig.~\ref{fig_allerrors}(a). 

We also include recovery error of the probability matrices in Fig.~\ref{fig_allerrors}(b), which demonstrates that joint inference of the networks and graphon in (4) can achieve significantly more accurate probability matrix estimates than joint inference of the networks and the probability matrices in (3). 
Joint graphon estimation not only assumes the correct signal model, where nodes across networks are not aligned, but also ensures that closer probability matrix entries are more similar, in accordance with the smooth graphon assumption.

Fig.~\ref{fig_t_figs} demonstrates the influence of the smooth graphon assumption on the estimated probability matrices, where we show a comparison of the inferred probability matrices from our proposed methods. 
For joint network and probability matrix inference (3), no dependence is assumed between the edge probabilities in the probability matrices, whereas joint network and graph inference (4) relates edges between nodes based on their assignments in the latent sample space.

Finally, graphon recovery error is shown in Fig.~\ref{fig_allerrors}(c). 
The joint inference method augmented with graphon estimation in red assumes that nodes are aligned across networks, an incorrect assumption given that networks are sampled from different latent point sets.
Thus, the augmented separate inference method exhibits superior graphon estimates.

\vspace{.1in}
\noindent{\bf Estimation of weighted networks.}
While our method focuses on learning the support of the estimated networks, there are many examples of network applications that require additional features such as edge weights.  
To this end, we combine our proposed method in a two-phase approach with another process that obtains node or edge features from network structure.
Indeed, graph-based learning is often approached in three steps: obtaining the network structure, assigning weights to the edges, and performing downstream learning tasks~\cite{2020_vanengelen_survey}.

We include results in Fig.~\ref{fig_edge_weights} using this two-phase approach, where we first estimate the network structures via our proposed method, then we obtain edge weights via similarity-based weights.
In particular, we apply the commonly used Gaussian edge weighting scheme~\cite{2020_vanengelen_survey}.
For two nodes $i$ and $j$ in the $k$-th network $\mathcal{G}^{(k)}$, if the edge $(i,j)$ exists, then we assign the edge weight ${W}_{ij}^{(k)}$ as
\begin{equation*}
W_{ij}^{(k)} = \exp\left\{
    \frac{-\|\mathbf{X}^{(k)}_i-\mathbf{X}^{(k)}_j\|_2^2}{2\sigma^2}
    \right\},
\end{equation*}
where $\sigma^2$ is the variance of the Gaussian kernel, and $\mathbf{X}_i^{(k)}$ denotes the vector of all graph signal values at the the $i$-th node of the $k$-th network.

Even when the underlying networks are weighted, implementing joint estimation of the probability matrix proves effective at improving graph estimation.
The combination of inferring network connectivity and estimating edge weights is not only empirically feasible but also straightforward in implementation, as there exist several strategies to obtain graph characteristics given network structure~\cite{2020_vanengelen_survey}.